\documentclass[journal]{IEEEtran}

\usepackage{amssymb}
\usepackage{amsthm}

\usepackage[ruled,vlined]{algorithm2e}
\usepackage{algorithmic}
\usepackage{graphicx}
\usepackage{epsfig}
\usepackage{amsmath}
\usepackage{lscape}
\usepackage{tabularx}
\usepackage{color, soul}
\usepackage{makecell}
\usepackage[table]{xcolor}
\usepackage{colortbl}
\usepackage{rotating}
\usepackage{mathtools}
\usepackage{hyperref}
\usepackage[left]{lineno}

\DeclarePairedDelimiter\ceil{\lceil}{\rceil}

\ifCLASSINFOpdf
\else
\fi
\begin{document}

%\linenumbers

\title{A Fully-autonomous Framework of Unmanned Surface Vehicles in Maritime Environments using Gaussian Process Motion Planning}

\author{Jiawei~Meng$^{1}$,~\IEEEmembership{Student Member,~IEEE,}
        Ankita~Humne$^{2}$,~\IEEEmembership{Student Member,~IEEE,}
        Richard~Bucknall$^{1}$,~\IEEEmembership{Member,~IEEE},
        Brendan~Englot$^{3}$,~\IEEEmembership{Senior Member,~IEEE}
        and Yuanchang~Liu$^{1, *}$,~\IEEEmembership{Member,~IEEE}
\thanks{J. Meng$^{1}$, Y. Liu$^{1,*}$ and R. Bucknall$^{1}$ are with the Department of Mechanical Engineering, University College London, Torrington Place, London WC1E 7JE, UK (corresponding author: Yuanchang Liu, yuanchang.liu@ucl.ac.uk, tel: +44 (0)20 7679 7062). A. Humne$^{2}$ is with the Department of Microtechnique (Robotics), EPFL, Switzerland. B. Englot$^{3}$ is with the Department of Mechanical Engineering, Stevens Institute of Technology, Hoboken, NJ, USA.
}
}

\maketitle

\begin{abstract}
 Unmanned surface vehicles (USVs) are of increasing importance to a growing number of sectors in the maritime industry, including offshore exploration, marine transportation and defence operations. A major factor in the growth in use and deployment of USVs is the increased operational flexibility that is offered through use of \textcolor{black}{optimised motion planners} that generate optimised trajectories. Unlike path planning in terrestrial environments, planning in the maritime environment is more demanding as there is need to assure mitigating action is taken against the significant, random and often unpredictable environmental influences from winds and ocean currents. With the focus of these necessary requirements as the main basis of motivation, this paper proposes a novel motion planner, denoted as \textcolor{black}{Gaussian process motion planning 2 star (}GPMP2*\textcolor{black}{)}, extending the application scope of the fundamental \textcolor{black}{Gaussian process-based (}GP-based\textcolor{black}{)} motion planner, \textcolor{black}{Gaussian process motion planning 2 (}GPMP2\textcolor{black}{)}, into complex maritime environments. An interpolation strategy based on Monte-Carlo stochasticity has been innovatively added to GPMP2* to produce a new algorithm named GPMP2* with Monte-Carlo stochasticity (MC-GPMP2*), which can increase the diversity of the paths generated. In parallel with algorithm design, a \textcolor{black}{Robotic Operating System} (ROS) based fully-autonomous framework for an advanced USV, the \textcolor{black}{Wave Adaptive Modular Vessel 20} \textcolor{black}{(}WAM-V 20\textcolor{black}{)}, has been proposed. The practicability of the proposed motion planner as well as the fully-autonomous framework have been functionally validated in a simulated inspection missions for an offshore wind farm in ROS.
\end{abstract}


% \fi
\IEEEpeerreviewmaketitle

\section{Introduction}
 The planning of trajectories in complex maritime environments plays a critical role in developing autonomous maritime platforms such as USVs. The paths generated for operations in maritime environments should not only ensure the success of a mission but, wherever and whenever possible, actively try to minimise the energy consumption during a voyage. Even with growing recognition of the importance of motion planning algorithms for USVs, two major challenges have largely hindered their progress of development including: 1) the majority of mainstream motion planning algorithms do not encompass proper consideration of the environmental impacts such as winds and surface currents and 2) among the minority of algorithms that do take these environmental characteristics into account, important metrics including the computation time and path quality are not up to that minimum standard of quality required for practical applications. \textcolor{black}{These aforementioned challenges have been addressed to some extent in the past few years, but there is a need to further optimise the solutions.}
 
 \begin{figure}[t!]
 \centering
 \includegraphics[width=1.0\linewidth]{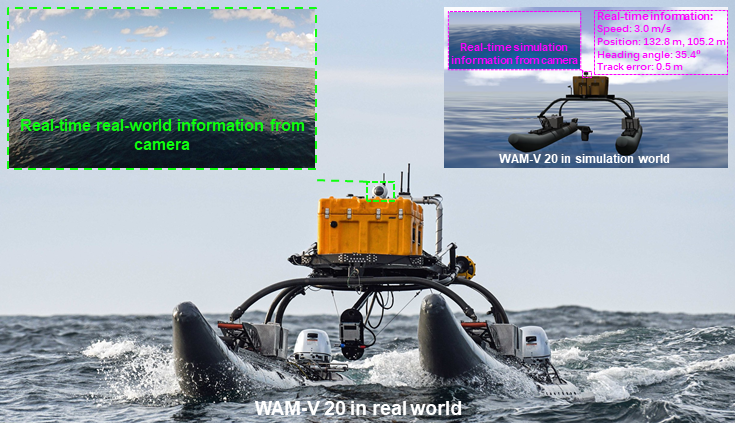}
 \centering
 \caption{A demonstration of WAM-V 20 USV in the real world and the virtual maritime scenario. The virtual maritime scenario is highly similar to the real world, where the real-time camera information, position, speed, heading angle and track error can be measured.}
 \label{WAM-V_in_Gazebo_and_real_world} 
 \end{figure}
 
 Current existing mainstream motion planning algorithms can be divided into four categories: 1) grid-based algorithms \cite{dijkstra1959note, hart1968formal, stentz1997optimal}, 2) sampling-based algorithms \cite{kavraki1996probabilistic, lavalle1998rapidly, 8665162}, 3) potential field algorithms \cite{khatib1986real, petres2005underwater} and \textcolor{black}{4)} intelligent algorithms \cite{dorigo1991distributed, whitley1994genetic, mahmoudzadeh2016novel}, variations of which have been applied across different robotic domains. \textcolor{black}{All the aforementioned algorithms have been developed over many years and have made an incredible contribution to robotic motion planning problems. Nevertheless, these algorithms have some drawbacks and cannot fully meet the requirements for motion planning in practical application scenarios.} Grid-based algorithms require a post-processing path smoothing procedure to satisfy the non-holonomic constraints of vehicles \cite{petres2007path}. Sampling-based and intelligent algorithms might require an extremely long computation time to ensure convergence, otherwise the distance and smoothness of the paths can not be guaranteed \cite{lolla2015path, goldberg1988genetic}. Potential field algorithms suffer from the limitations of local minima and require additional strategies to avoid this issue \cite{garrido2008exploration}. \textcolor{black}{Meanwhile, these motion planning algorithms are not designed for maritime environments with time-varying ocean currents. Another perspective of categorising different motion planning algorithms can be found in \cite{singh2018optimal}.} 
 
 To address the problems \textcolor{black}{in practical application scenarios}, trajectory optimisation algorithms have been proposed in recent years \cite{mukadam2016gaussian, barfoot2014batch, yan2017incremental, dong2016motion, mukadam2018continuous}. One of them is the \textcolor{black}{GP-based} motion planning algorithm \cite{mukadam2016gaussian,mukadam2018continuous} that represents trajectories as samples from Gaussian processes in the continuous-time domain and optimises them via probabilistic inference. This novel motion planning paradigm brings two significant benefits: 1) the capability of smoothing the path in line with the planning process based on the specification of the system's dynamic models and 2) the superiority in convergence speed through the employment of a fast-updating inference tool such as a factor graph \cite{dellaert2012factor}. However, there are still some constraints when it comes to implementing trajectory optimisation algorithms in maritime environments, and the issues of integrating characteristics of maritime environments such ocean currents and avoiding dense obstacles remains especially challenging.
 
 Another research bottleneck for USV development is the lack of high-fidelity environments. Fig. \ref{WAM-V_in_Gazebo_and_real_world} compares a typical catamaran, the WAM-V 20 USV, in real world and virtual maritime scenarios. In this high-fidelity virtual maritime scenario, physical fidelity and visual realism with real-time execution requirements are well-balanced. In general, establishing practical experimental platforms would be expensive. By developing high-fidelity simulation environments, validating the newly proposed motion planning, control and any other algorithms can be conducted in an efficient and low-cost manner.   

 In fact, simulations with a sufficient level of fidelity have been gradually adopted for USV platforms. Game engines such as Unity \cite{unity} and Unreal Engine \cite{unreal_engine} can present a vivid virtual world, which might be suitable simulation platforms for motion planning and control algorithms. However, most of them do not have a dedicated support for robotics and the hardware requirements for running these game engines are usually difficult to satisfy. In 2002, an open-source simulation platform designed for supporting various indoor and outdoor robotic applications was proposed, namely the Gazebo \cite{gazebo}. Specifically, it delivers the following benefits that made it become the most popular simulation platform among robotic researchers: 1) it supports the use of different physics engines to simulate collision, contact and reaction forces among rigid bodies, 2) its sensor libraries are progressive due to the open source facility and 3) it supports for robotics middle-ware based upon a well-developed messaging system. 

 Nevertheless, most of the simulators based on the fundamental structure of Gazebo are designed for terrestrial, aerial and space robots \cite{aguero2015inside, allan2019planetary, furrer2016rotors}. To address this deficiency and provide a standard simulator for the development and testing of algorithms for USVs, the Virtual RobotX simulator (VRX) was proposed in 2019. VRX is a Gazebo-based simulator capable of simulating the behaviour of USVs in complex maritime circumstances with waves and buoyancy conditions \cite{vrx}. Also, a mainstream catamaran (WAM-V 20 USV) model is provided in VRX with an easy-to-access interface to any self-designed autopilot. There is, however, a lack of a fully-autonomous navigation system in VRX, especially a system integrating both motion planning capability and autopilot.
 
 \textcolor{black}{To bridge these research gaps, this paper has specifically focused on developing a new motion planning paradigm for USVs with the main contributions summarised as follows:} 
 \begin{itemize}
     \item \textcolor{black}{A new GP-based motion planning algorithm, named as MC-GPMP2*, has been developed by integrating a Monte-Carlo stochasticity to enable an improved collision avoidance capability.}
     \item \textcolor{black}{A fully-autonomous framework for USVs has been designed for the VRX simulator.}
     \item \textcolor{black}{Enriched high-fidelity tests have been carried out in ROS to simulate offshore wind farm operations using USVs, where the superiority of the proposed motion planning algorithms is properly demonstrated.}
 \end{itemize}

 The rest of the paper is organised as follows. Section 2 formulates the problem and discusses the mathematical model of the conventional GP-based motion planning algorithm in various complex environments. Section 3 describes the Monte-Carlo sampling and introduces it into our motion planning algorithm. Section 4 presents the modelling and control of the WAM-V 20 USV in ROS. Section 5 demonstrates the proposed path planner's simulation results and then compares them with the results obtained from a series of mainstream motion planning algorithms. Section 6 demonstrates the practical performance of the proposed path planning algorithm and autopilot in ROS, followed by the conclusion and indications for future work in Section 7.

\section{GP-based Motion Planning in Various Complex Environments}
\label{Guassian_process_motion_planner_2_star}
 This section explains the GP-based motion planning algorithms in general and proposes a new method named the Gaussian process motion planner 2 star (GPMP2$^*$), which will be developed and applied to motion planning for autonomous vehicles such as USVs and \textcolor{black}{unmanned underwater vehicles (}UUVs\textcolor{black}{)}.
 
\subsection{Problem formulation as trajectory optimisation}
\label{problem_formulation}
 GP-based motion planning algorithms can be applied to solve the problem of trajectory optimisation, i.e. employing Gaussian Processes to optimise trajectories in an efficient manner. Formally, the trajectory optimisation aims to determine the best trajectory from all feasible trajectories while satisfying any user defined constraints and minimising any user prioritised costs \cite{ko2009gp, gammell2015batch, yan2017incremental}. By considering a trajectory as a function of continuous time $t$, such an optimisation process can be formulated as the standard form of an optimisation problem with continuous variables as:
 
 \begin{equation}
 \begin{aligned}
 & {\text{minimise}}
 & & F[\theta(t)] \\
 & \text{subject to}
 & & G_{i}[\theta(t)] \leq 0, \; i = 1, \ldots, m_{ieq}\\
 & & & H_{i}[\theta(t)] = 0, \; i = 1, \ldots, m_{eq}.
 \end{aligned}
 \label{trajectory_optimisation}
 \end{equation}
 where $\theta(t)$ is a continuous-time trajectory function mapping a specific moment $t$ to a specific robot state $\theta$. $F[\theta(t)]$ is an objective function to find the best trajectory by minimising the higher-order derivatives of robot states (such as velocity and acceleration) and collision costs. $G_{i}[\theta(t)]$ is a task-dependent inequality constraint function and $H_{i}[\theta(t)]$ is a task-dependent equality constraint function that contain the desired start and goal robot states with specified configurations.
 
 As stated in \cite{mukadam2016gaussian, meng2022anisotropic}, by properly allocating the parameters of low-resolution states \textcolor{black}{with relatively large sample interval $\Delta t$} (defined as support states) and interpolating high-resolution states \textcolor{black}{with relatively small sample interval $\Delta \tau$} (defined as interpolated states), the computational cost of Gaussian Processes can be efficiently reduced and a continuous-time trajectory function represented by a Gaussian Process can be shown as:
 
 \begin{equation}
 \theta(t)\sim\mathcal{GP}(\mu(t),K(t,t')), 
 \end{equation}
 
 \noindent where $\mu(t)$ is a vector-valued mean function and $K(t,t')$ is a matrix-valued covariance function.
 
 For the given optimization problem, the objective function  is given as:
 
 \begin{equation}
     F[\theta(t)] = F_{gp}[\theta(t)] + \textcolor{black}{\omega_{1}}F_{obs}[\theta(t)] + \textcolor{black}{\omega_{2}}F_{env}[\theta(t)], 
     \label{objective_function}
 \end{equation}
where $F_{gp}[\theta(t)]$ is the GP prior cost, $F_{obs}[\theta(t)]$ is the obstacle collision cost and $F_{env}[\theta(t)]$ is the environment characteristic cost. \textcolor{black}{$\omega_{1}$} and \textcolor{black}{$\omega_{2}$} are the weight coefficients given to these costs. At this juncture we specifically highlight the inclusion of the environment cost ($F_{env}[\theta(t)]$) as it is of particular importance when considering marine vehicles. For other types of vehicles, costs can be adjusted as required.
 
\subsection{Motion planning as probabilistic inference}
\textcolor{black}{From another perspective, GP-based motion planning algorithms can also be viewed as probabilistic inference problems, where Bayesian inference is applied to find the optimal trajectory.} \textcolor{black}{The detailed explanation of using Bayesian inference to solve the trajectory optimisation problem in (\ref{trajectory_optimisation}) can be found in \cite{dong2016motion}.} In this subsection, we summarise the work in \cite{dong2016motion} and extend it to the more general case that includes multiple planning constraints. 

By exploiting the sparsity of the underlying problem, probabilistic inference, such as Bayesian inference, is effective for solving optimisation problems and the optimisation problem in (\ref{trajectory_optimisation}) can be converted into the following Bayesian inference:
 
 \begin{align}
    \theta^* = \mathop{\arg\max}_{\theta} \; p(\theta)l(\theta; e),
 \label{maximum_a_posterior}
 \end{align}
 
\noindent where $p(\theta)$ represents the \textit{GP prior} that encourages smoothness of trajectory and $l(\theta; e)$ represents a \textit{likelihood}. More specifically, the \textit{GP prior} distribution is given in terms of the mean $\mu$ and covariance $K$:
 
 \begin{equation}
    p(\theta) \propto \textcolor{black}{\exp}\{ -\frac{1}{2}||\theta - \mu||^{2}_{K}\},
 \label{prior_distribution}
 \end{equation}
 
 \noindent whereupon the \textit{GP prior} cost in (\ref{objective_function}) is given as the negative natural logarithm of the prior distribution:
 
 \begin{equation}
    F_{gp}[\theta(t)] = F_{gp}[\theta] = \frac{1}{2}||\theta - \mu||^{2}_{K}.
 \label{GP_prior_cost}
 \end{equation}
 
The \textit{likelihood} in the above Bayesian inference can be viewed as a combination of different categories of likelihoods such as the obstacle collision likelihood and the environment characteristic likelihood, thereby it is named as the \textit{combined likelihood}. Moreover, it is worth noting that the distribution of the \textit{combined likelihood} can be written into a product of the distributions from all the subcategory likelihoods using the features of the exponential distribution:
 
 \begin{align}
     l(\theta; e) = \underbrace{\textcolor{black}{\exp}\{ -\frac{1}{2}||g_{1}(\theta)||^{2}_{\Sigma_{obs}}\}}_{l(\theta; e_{obs})} \cdot \underbrace{\textcolor{black}{\exp}\{ -\frac{1}{2}||g_{2}(\theta)||^{2}_{\Sigma_{env}}\}}_{l(\theta; e_{env})}\\
     = \textcolor{black}{\exp}\{ -\frac{1}{2}||g_{1}(\theta)||^{2}_{\Sigma_{obs}}-\frac{1}{2}||g_{2}(\theta)||^{2}_{\Sigma_{env}}\}
  \label{combined_likelihood_cost_distribution}
 \end{align}

 $\Sigma_{obs}$ and $\Sigma_{env}$ are the diagonal covariance matrices with regard to collision and environmental characteristics: 
 
 \begin{equation}
    \textcolor{black}{\Sigma_{\textit{obs}(\textit{env})} = \rm{diag}[\sigma_{obs(env)}],}
 \end{equation}
 
 \noindent \textcolor{black}{where} $\sigma_{\textit{obs}}$ and $\sigma_{\textit{env}}$ are the weighting coefficients with regard to collision and environment characteristics. $g_{1}(\theta)$ and $g_{2}(\theta)$ are defined as a vector-valued obstacle cost function and a vector-valued environment characteristic cost function. More specifically, the definition of $g_{1}(\theta)$ is given as:
 
 \begin{align}
         g_{1}(\theta_{i}) = [c(d(x(\theta_{i}, S_{j})))]_{1 \leq j \leq M},
 \label{workspace_cost_function}
 \end{align}
 
 \begin{figure}[t!]
 \centering
 \includegraphics[width=1.0\linewidth]{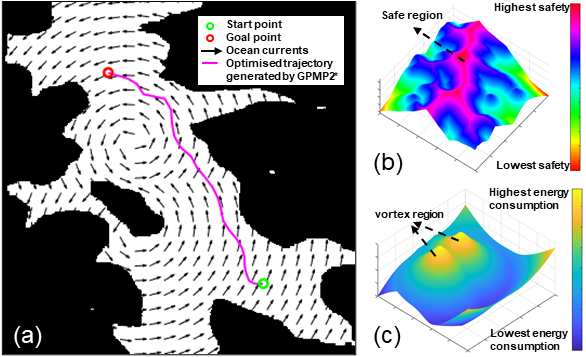}
 \centering
 \caption{An example of the proposed GPMP2* motion planning algorithm: (a) demonstrates the optimised trajectory generated by GPMP2*, (b) demonstrates the signed distance field generated by the obstacle collision likelihood function and (c) demonstrates the environment characteristic field generated by the environment characteristic likelihood function.}
 \label{GPMP2_star} 
 \end{figure}
 
 \noindent where $c(\cdot): \mathbb{R}^{n} \rightarrow \mathbb{R}$ is the workspace cost function that penalises the set of points $B \subset \mathbb{R}^{n}$ on the robot body when they are in or around an obstacle, $d(\cdot): \mathbb{R}^{n} \rightarrow \mathbb{R}$ is the signed distance function that calculates the signed distance of the point, $x$ is the forward kinematics function, $S_{j}$ is the sphere on the robot model and $M$ is the number of spheres that represents the robot model. An example of a constructed signed distance field is graphically shown in Fig. \ref{GPMP2_star} (b), and the obstacle collision cost in (\ref{objective_function}) can be given as:
 
 \begin{equation}
    F_{obs}[\theta(t)] = \int^{t_{N}}_{t_{0}}\int_{B}c(x\textcolor{black}{(\theta(t), u)})||\dot{x}\textcolor{black}{(\theta(t), u)}||dudt.
 \label{obstacle_collision_cost}
 \end{equation}
 
 \noindent where $u$ represents the known system control input. Also, the definition of $g_{2}(\theta)$ is given as:
 
 \begin{equation}
         g_{2}(\theta_{i}) = [e(x(\theta_{i}, S_{j}))]_{1 \leq j \leq M},
 \label{environment_compensation_function}
 \end{equation}
 
\noindent where $e(\cdot): \mathbb{R}^{n} \rightarrow \mathbb{R}$ is the environment compensation function that integrates the relevant environment characteristics such as surface wind and ocean currents on the set of points $B \subset \mathbb{R}^{n}$ on the robot body.

The environment compensation function is defined as a metric calculated using an anisotropic fast marching algorithm as stated in \cite{song2017multi, meng2022anisotropic}. Such a metric can measure the energy consumption rate at each pixel so that trajectories can be generated to avoid high energy consumption regions (the bright regions in Fig. \ref{GPMP2_star} (c)). The environment information is simulated as a vortex function in this work but any real-time statistical data can also be extracted and used as stated in \cite{lolla2015path}. The environment characteristic cost in (\ref{objective_function}) can therefore be given as:
 
 \begin{align}
    F_{env}[\theta(t)] = \int^{t_{N}}_{t_{0}}\int_{B}e(x\textcolor{black}{(\theta(t), u)})||\dot{x}\textcolor{black}{(\theta(t), u)}||dudt,
 \label{environment_characteristic_cost}
 \end{align}
 
 \noindent where $u$ represents the known system control input.
 
 Now we can rewrite the Bayesian inference in (\ref{maximum_a_posterior}) into the following form on the basis of the information provided by (\ref{prior_distribution}) and (\ref{combined_likelihood_cost_distribution}): 
 
 \begin{align}
        \theta^* & = \mathop{\arg\max}_{\theta} \; p(\theta)l(\theta; e)\\
        & = \mathop{\arg\max}_{\theta} \; \{-\textcolor{black}{\log}(p(\theta)l(\theta; e))\}\\
        & = \mathop{\arg\max}_{\theta} \; \{\frac{1}{2}||\theta - \mu||^{2}_{K}+\frac{1}{2}||g_{1}(\theta)||^{2}_{\Sigma_{obs}}+\frac{1}{2}||g_{2}(\theta)||^{2}_{\Sigma_{env}}\}.
 \end{align}

 Similarly, we can rewrite the objective function in (\ref{objective_function}) into the following form on the basis of the information provided by (\ref{GP_prior_cost}), (\ref{obstacle_collision_cost}) and (\ref{environment_characteristic_cost}):

 \begin{equation}
 \begin{gathered}
     F[\theta(t)] = \frac{1}{2}||\theta - \mu||^{2}_{K} + \lambda_{1}\int^{t_{N}}_{t_{0}}\int_{B}c(x\textcolor{black}{(\theta(t), u)})||\dot{x}\textcolor{black}{(\theta(t), u)}||dudt\\+\lambda_{2}\int^{t_{N}}_{t_{0}}\int_{B}e(x\textcolor{black}{(\theta(t), u)})||\dot{x}\textcolor{black}{(\theta(t), u)}||dudt, 
 \label{objective_function_new}
 \end{gathered}
 \end{equation}
 
 \noindent where $\lambda_{1}$ and $\lambda_{2}$ correspond to $\sigma_{obs}$ and $\sigma_{env}$, respectively.
 
 A notable advantage of the proposed motion planning algorithm is that when multiple environment constraints need to be considered simultaneously, these constraints can be formulated as various subclass environment characteristic likelihoods. By taking advantage of the features of exponential distributions, subclass likelihoods can be further integrated into a superclass environment characteristic likelihood to enable a fast summation of constraints. To gain a more intuitive understanding of the feasibility of the proposed motion planning algorithm, Fig. \ref{GPMP2_star} demonstrates an example of how the GP-based motion planner can be used to avoid obstacles as well as vortexes. Also, the proposed method can be used in either \textcolor{black}{2-dimensional} (2D) or \textcolor{black}{3-dimensional} (3D) environments. Overall, any type of GP-based motion planning method that incorporates the characteristics of the environment through adding corresponding likelihood to probabilistic inference can be viewed as GPMP2*.
 
\section{GP-based Motion Planning with Incremental Optimisation Characteristics}
This section provides \textcolor{black}{detail} of the proposed MC-GPMP2* algorithm. The Monte-Carlo sampling based, obstacle space estimation is first introduced and this is then followed by the details of sampling point interpolation and incremental inference. 

 \subsection{Obstacle space estimation using Monte-Carlo sampling}
Monte-Carlo sampling is a highly efficient statistical method to determine the approximate solution of many quantitative numerical problems. It can reduce the computation time when there is a relatively high complexity in sampling space \cite{metropolis1949monte, eckhardt1987stan}. In this work, this sampling method is used to estimate the ratio of the obstacle space to the entire sampling space, especially when the obstacle space has a relatively irregular shape. Algorithm \ref{alg:1} demonstrates the specific procedure of the estimation, where the random sample point is generated from a continuous uniform distribution:
 
 %% Pseudocode of Monte-Carlo sampling
 \begin{algorithm}[t!]
 \small
 \SetAlgoLined
 \SetAlCapNameFnt{\normalsize}
 \SetAlCapFnt{\normalsize}
 \textbf{Input:} 3-dimensional sampling space $\mathcal{X}_{x, y, z}$ and the total number of samples $N_{spl}$\\
 \For{$i=1,2,\ldots,N_{spl}$}
        {
            Generate a random sample point inside the 3-dimensional sampling space $\textit{X}_{rand} \leftarrow$ $\texttt{Sample}$($\mathcal{X}_{x, y, z}$);\\
            \If{($\texttt{CollisionFree}$($\textit{X}_{rand}, \mathcal{X}_{x, y, z}$) == TRUE)}
            {
                Accept the random sample point $X_{rand}$ by increasing the accepted sample number $N_{ac} = N_{ac} + 1$\\
            }
            \If{($\texttt{CollisionFree}$($\textit{X}_{rand}, \mathcal{X}_{x, y, z}$) == FALSE)}
            {
                Reject the random sample point $X_{rand}$ by maintaining the previous accepted sample number $N_{ac} = N_{ac}$\\
            }
		}
		Compute the ratio of the obstacle space to the entire sampling space through $P_{obs} = \frac{N_{ac}}{N_{spl}}$\\
 \textbf{Output:} Obstacle space proportion $P_{obs}$\\
 \textbf{Notes:} The pixels inside the obstacle space are '1' and the pixels outside the obstacle space are '0'. \textcolor{black}{$\texttt{CollisionFree}$($\textit{X}_{rand}, \mathcal{X}_{x, y, z}$ is a function to check whether the random generated node $X_{rand}$ is inside the obstacle space or not.}\\
 \caption{\small Obstacle Space Estimation using the Monte-Carlo Sampling ($\texttt{MC-EstimateObstacleSpace}$)}
 \label{alg:1}
 \end{algorithm}
 
 \begin{equation}
 (x, y, z)\sim\mathcal{U}(a, b), 
 \end{equation}
 
 \noindent where $a = (a_{1}, a_{2}, a_{3})^{T}$ is a vector-valued lower bound function representing the lower bounds of the 3-dimensional sampling space, $b = (b_{1}, b_{2}, b_{3})^{T}$ is a vector-valued upper bound function indicating the upper bounds of the sampling space. The probability density function of the continuous uniform distribution at any point $(x, y, z) \in \mathbb{R}^{3}$ inside the sampling space can be given as:
 
 \begin{equation}
 f(\ceil{x},\ceil{y},\ceil{z}) = \frac{1}{\prod_{i=1}^{3} (b_{i}-a_{i})},
 \end{equation}
 
 \noindent where $\ceil{\cdot}$ is a \textcolor{black}{ceiling} function used to round-up to the nearest integer and the volume of the entire sampling space can be represented by $\prod_{i=1}^{3} (b_{i}-a_{i})$.
     
 In Algorithm \ref{alg:1}, based on the law of large numbers \cite{lln, grimmett2020probability, durrett2019probability}, the accuracy of the estimation of obstacle space gradually increases as the number of samples increases. Its convergence rate is O$({\frac{1}{\sqrt{N}}})$, which means that quadrupling the total number of samples reduces the algorithm's error by half, regardless of the dimensions of the sampling space \cite{asmussen2007stochastic}. Therefore, Algorithm \ref{alg:1} can provide a reasonably accurate result once the total number of samples exceeds a specific threshold. When the total number of samples in Algorithm \ref{alg:1} equals the total number of pixels on the map, the sampling algorithm becomes a traversal algorithm and generates a result with an accuracy that can be considered absolute. 
 
 In general, using GP-based motion planning in conjunction with Algorithm \ref{alg:1} \textcolor{black}{to construct a modified motion planning algorithm }can offer two notable advantages:
 \begin{itemize}
     \item Shortening the execution time of GP-based motion planning, especially for high-dimensional problems;
     \item Shortening the path length and improving the path quality by enhancing the diversity of the generated trajectory.
 \end{itemize}

\subsection{Monte-Carlo based GP interpolation}
As mentioned in Section. \ref{Guassian_process_motion_planner_2_star}, apart from the support states, a major benefit of using GPs is the facility to query the planned state at any moment of interest. In addition, according to \cite{dong2016motion}, trajectories generated by a GP-based motion planner can be fine-tuned by increasing the number of states. Therefore, to facilitate obstacle avoidance, in this paper, it is proposed to interpolate additional states between two support states according to the obstacle estimation of Monte-Carlo. 

Similar to previous research \cite{dong2016motion, yan2017incremental, mukadam2016gaussian, mukadam2018continuous}, a linear time-varying stochastic differential equation (LTV-SDE) is adopted to represent the motion model as:
 
 \begin{equation}
     \dot{\theta}(t) = A(t)\theta(t) + u(t) + F(t)w(t).
 \label{LTV_SDE}
 \end{equation}
where $A(t)$ and $F(t)$ are time-varying matrices of the system, $u(t)$ is the control input and $w(t)$ is the white process noise represented as:

\begin{equation}
    w(t)\sim\mathcal{GP}(0,Q_c\delta(t,t')), 
\end{equation}
where $Q_c$ is the power-spectral density matrix and $\delta(t,t')$ is the Dirac delta function. Based on (\ref{LTV_SDE}), a queried/interpolated state $\theta(\tau)$ at $\tau \in [t_{i}, t_{i+1}]$ is a function only of its neighboring state as (the detailed proof of this is presented in \cite{dong2016motion, yan2017incremental, mukadam2016gaussian, mukadam2018continuous}):
 
 \begin{equation}
    \theta(\tau) = \widetilde{\mu}(\tau) + \Lambda(\tau)(\theta_{i} - \widetilde{\mu}_{i}) + \Psi(\tau)(\theta_{i+1} - \widetilde{\mu}_{i+1}),
 \label{fast_GP_interpolation}
 \end{equation}
 where 
 \begin{equation}
    \begin{array}{cc}
         \Lambda(\tau) = \Phi(\tau,t_i) - \Psi(\tau)\Phi(t_{i+1},t_i),&  \\
         \Psi(\tau) = Q_{i,\tau}\Phi(t_{i+1},\tau)^TQ_{i,i+1}^{-1}& ,
    \end{array}
 \end{equation}

 \noindent where $\Phi(*,*)$ is the state transition matrix and $Q_{a,b}$ is:
 
 \begin{equation}
    Q_{a,b} = \int_{t_a}^{t_b}\Phi(b,s)F(s)Q_{c}F(s)^T\Phi(b,s)^T\textit{ds}.
 \label{Q_a_b}
 \end{equation}

 %%Building Factor Graph with the Monte-Carlo Stochasticity
 \begin{algorithm}[t!]
 \small
 \SetAlgoLined
 \SetAlCapNameFnt{\normalsize}
 \SetAlCapFnt{\normalsize}
 %\textbf{Input:} Total number of sub-searching regions $N$ and the maximum sampling time $T_{max}$\\
 \textbf{Input:} Total number of sub-searching regions $N$\\
 Add $\texttt{Prior Factor}$\\
 %Compute the low-resolution sample interval $\Delta{T}=T_{max}/N$\\
 \For{$i=1,2,\ldots,N$}
 {
        Add $\texttt{Obstacle Factor}$ and $\texttt{Environment Factor}$\\
		Compute total number of sample points in the low-resolution region $N_{j}$:\\
		$P_{obs} \leftarrow \texttt{MC-EstimateObstacleSpace}$\\
		$N_{j}=\lambda \cdot P_{obs}$\\
		%Compute the high-resolution sample interval $\Delta{\tau_{j}} = \frac{\Delta T}{N_{j}}$\\
		\For {$j=1,2,\ldots,N_{j}$}
		{
			%Perform GP interpolation by using (\ref{dynamic_fast_GP_interpolation})\\
			Add $\texttt{GP Prior Factor}$, $\texttt{Interpolated Obstacle Factor}$ and $\texttt{Interpolated Environment Factor}$\\
		}
 }
 Add $\texttt{Prior Factor}$\\
 \textbf{Output:} Factor graph $G_{mc}$\\
 \textbf{Notes:} $\lambda$ represents a self-defined scaling term.\\
 \caption{\small Building Factor Graph with the Monte-Carlo Stochasticity ($\texttt{MC-BuildFactorGraph}$)}
 \label{alg:2}
 \end{algorithm}
 
 \begin{figure}[t!]
 \centering
 \includegraphics[width=1.0\linewidth]{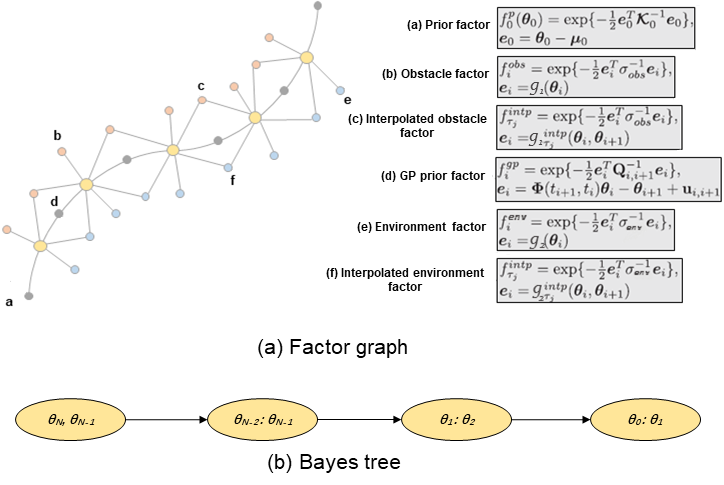}
 \centering
 \caption{A demonstration of the factor graph and Bayes tree in our problem: (a) illustrates the factor graph, containing six categories of factors including Prior factor, GP prior factor, Obstacle factor, Interpolated obstacle factor, Environment factor and Interpolated environment factor, (b) illustrates the Bayes tree, indicating the conditional dependencies between various states.}
 \label{factor_graph&bayes_tree} 
 \end{figure}

In general, (\ref{fast_GP_interpolation}) can be used to interpolate a series of dense states to facilitate the generation of collision-free trajectories while keeping a relatively small number of support states to maintain low computational cost. As stated previously, the strategy of interpolating dense states is lacking in the previous research \cite{dong2016motion, yan2017incremental, mukadam2016gaussian, mukadam2018continuous}. Therefore, it is proposed that the number of interpolated states should be determined by the proportion of the obstacle space relative to the whole region space \textcolor{black}{($P_{obs}$)}, and such a proportion can be quickly estimated by Monte-Carlo sampling as described in Algorithm \ref{alg:1}. In addition, the Monte-Carlo stochasticity adds a variation to the number of interpolated states ($N_{j} = \frac{t_{i+1} - t_{i}}{\tau} = \lambda \cdot P_{obs}$) to test the optimal number of interpolated states incrementally, where $\lambda$ is a self-defined scaling term and $P_{obs}$ is computed by Algorithm \ref{alg:1}. \textcolor{black}{More specifically, $P_{obs}$ tends to increase as the volume of obstacles within a specific region increases, leading to a growth in the number of interpolated states of this region $N_{j}$. Interpolated states with relatively high densities can improve the performance of the motion planning algorithm on avoiding obstacles as well as smoothen the generated trajectory.}
 
 \subsection{Probabilistic inference using the factor graph}
 
Given the Markovian structure of the trajectory enabled by the \textcolor{black}{linear time-varying stochastic differential equation} (LVT-SDE) and the sparsity of the underlying problem, the posterior distribution (or the optimised trajectory) can be converted into a factor graph to perform inference incrementally. More specifically, the factor graph is a bipartite graph that can express any inference in a more intuitive graphical manner. It is bipartite as there are only two categories of nodes existing in the graph, i.e. variable nodes and factor nodes \cite{dellaert2012factor}. The factorisation of the posterior in our problem is formulated as:
 
 \begin{equation}
    p(\theta|e) \propto \prod^{M}_{m=1}f_{m}(\Theta_{m}),
 \label{factor_graph}
 \end{equation}
 
 \noindent where $f_{m}$ are factors on variable subset $\Theta_{m}$. 
 
 Then the factor graph can be converted into a Bayes tree based on the variable elimination process \cite{dellaert2006square, kaess2008isam, kaess2012isam2, kaess2011isam2}. The Bayes tree in our problem, as converted by (\ref{factor_graph}), is:
 
 \begin{equation}
    p(\Theta) = \prod^{}_{j}p(\theta_{j}|S_{j}),
 \label{Bayes_tree}
 \end{equation}
 
 \noindent where $\theta_{j}$ are the states and $S_{j}$ denotes the separator for state $\theta_{j}$ which is comprised of the nodes in the intersection of the state $\theta_{j}$ and its parent. 
 
 To gain a more intuitive understanding regarding the factor graph, a comprehensive structure illustrating how the different factors are integrated as well as converted into a Bayes tree for our problem is demonstrated in Fig. \ref{factor_graph&bayes_tree}. Furthermore, the specific process of building a factor graph with the Monte-Carlo stochasticity is detailed in Algorithm \ref{alg:2}.
 
  %%The Proposed Gaussian Process Motion Planner 2 star with the Monte-Carlo Stochasticity
 \begin{algorithm}[t!]
 \small
 \SetAlgoLined
 \SetAlCapNameFnt{\normalsize}
 \SetAlCapFnt{\normalsize}
 \textbf{Input:} Start state $\theta_{0}$, goal state $\theta_{N}$ and replanning iteration $N_{replan}$\\
 \textbf{Precompute} Signed distance field ($SDF$) and environment characteristic field ($ECF$)\\
 \For{$i=1,2,\ldots,N_{replan}$}
        {
            $G_{mc}$ $\leftarrow$ $\texttt{MC-BuildFactorGraph}$\\
            $\theta^{*}(i)$ $\leftarrow$ $\texttt{LM}$($\theta_{0}$, $\theta_{N}$, $G_{mc}$, $SDF$, $ECF$)\\
            $\theta^{*}$ $\leftarrow$ $\theta^{*}(i)$\\
            \If{\rm{\{}$L[\theta^{*}(i-1)] <= L[\theta^{*}(i)]$\} \rm{or} \{$\texttt{CollisionFree}$[$\theta^{*}(i)$] == FALSE\}}
            {
                $\theta^{*}$ = $\theta^{*}(i-1)$\\
            }
		}
 \textbf{Output:} Optimal path $\theta^{*}$\\
 \textbf{Notes:} $SDF$ is calculated by inputting the motion planning space into the workspace cost function. $ECF$ is calculated by inputting the motion planning space into the environment compensation function. $\texttt{LM}(\cdot)$ represents the Levenberg-Marquardt algorithm. $\texttt{L}(\cdot)$ represents the function to measure the total length of the generated path.\\
 \caption{\small Gaussian Process Motion Planner 2 star with the Monte-Carlo Stochasticity ($\texttt{MC-GPMP2*}$)}
 \label{alg:3}
 \end{algorithm}
 
 \subsection{Incremental optimising motion planning based on GPs}
 The Gaussian process motion planner 2 star with the Monte-Carlo stochasticity (MC-GPMP2*) is proposed in this subsection by integrating the aforementioned information. The pseudo-code of the proposed motion planner is detailed in Algorithm \ref{alg:3} with key information explained as: 
 \begin{itemize}
     \item First, the start state $\theta_{0}$, goal state $\theta_{N}$ and replanning iteration $N_{replan}$ are required as inputs.
     \item Next, the signed distance field is computed based on the obstacle cost function (as described in (\ref{workspace_cost_function})) and the environment characteristic field is computed based on the environment characteristic cost function (as described in (\ref{environment_compensation_function})), to construct the \textit{combined likelihood} (as described in (\ref{combined_likelihood_cost_distribution})).
     \item A factor graph with Monte-Carlo stochasticity is built based on the $\texttt{MC-EstimateObstacleSpace}$ (Algorithm \ref{alg:1}) and the $\texttt{MC-BuildFactorGraph}$ (Algorithm \ref{alg:2}) and then the optimal path $\theta^{*}$ is inferred based on the Levenberg-Marquardt algorithm \cite{levenberg1944method}. 
     \item The previous step is repeated several times based on the number of replanning iterations $N_{replan}$ required to optimise the path $\theta^{*}$.
 \end{itemize}
 
 \begin{figure}[t!]
 \centering
 \includegraphics[width=1.0\linewidth]{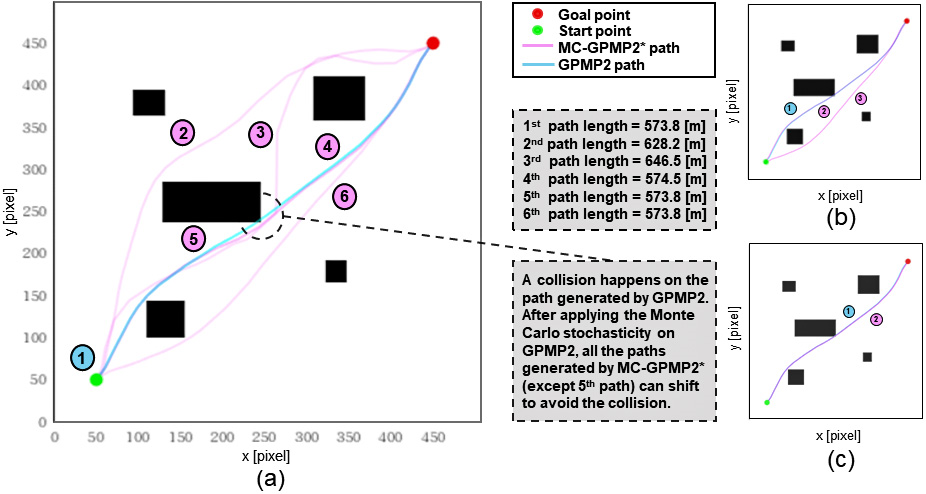}
 \centering
 \caption{A comparison of the paths generated by GPMP2 and MC-GPMP2*. GPMP2 generates a single solution, while MC-GPMP2* extends the form of this solution by adding randomness to the sampling process making the paths generated by MC-GPMP2* diversified. This characteristic provides extra solutions to a specified motion planning problem, i.e. increasing the probability of approaching a better path. From (a) - (c), the number of sample points increases gradually and the diversity of the paths generated by MC-GPMP2* decreases accordingly.}
 \label{comparison_paths} 
 \end{figure}
 
 To better understand the functionality of the proposed motion planner, a comparison of the paths generated by MC-GPMP2* and GPMP2 is presented in Fig. \ref{comparison_paths}. MC-GPMP2* generates relatively diversified paths when there are a relatively small number of sample points. Conversely, MC-GPMP2* generates a path with a high level of similarity compared with GPMP2 when there is a relatively larger number of sample points.
 
\section{WAM-V 20 USV modeling and control in ROS}
 In this section, detail will be provided regarding the proposed fully-autonomous navigation framework to navigate and control WAM-V 20 USVs in ROS. Overall, the proposed framework includes three major components: 1) {motion planner}, generating an optimised path according to obstacles and environment characteristics, 2) {navigation refinement system} to generate the USV heading angles needed to accurately track the paths and 3) {autopilot} adjusting the angle of deflection of rudders and the rotational speed of USVs to match the desired values.

\subsection{Mathematical modeling for WAM-V 20 USV}
\begin{figure}[t!]
 \centering
 \includegraphics[width=0.9\linewidth]{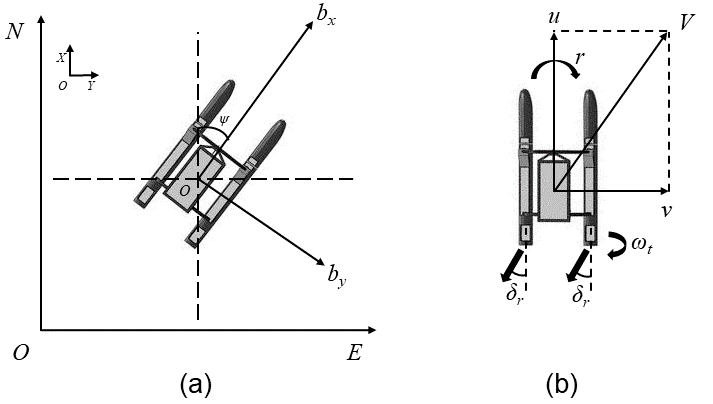}
 \centering
 \caption{Schematic depictions of the used catamaran: (a) shows the north-east-down reference frame $\textbf{N} = \{E, N\}$ and the frame attached to the USV platform $\textbf{B} = \{b_{x}, b_{y}\}$ and (b) shows the motion diagram of the USV, where $r$ is Z-axis angular velocity (or the USV yaw rate), $V$ is the net velocity of the USV, $u$ is the component of the net velocity on due north (or the USV surge velocity), $v$ is the component of the net velocity on due east (or the USV sway velocity), $\delta_{r}$ is the angle of deflection of rudders and $\omega_{t}$ is the rotational speed of the thrusters.}
 \label{schematic_depiction_usv} 
 \end{figure}

 The specifications of the catamaran that will be used are listed in Table \ref{Table:wamv}. This catamaran consists of a wave-adaptive structure and two air cushions with thrusters mounted at the back end of each cushion. The thrusters rotate around the Z axis simultaneously to supply different-oriented propulsion within the E-N plane as shown in Fig. \ref{schematic_depiction_usv}.
 
 The spatial position state of the catamaran $\dot{\overrightarrow{X}}$ is considered to be its 2D position $(E, N)$, heading angle $\psi$, sway velocity $v$, surge velocity $u$, yaw rate $r$, angle of deflection of rudders $\delta_{r}$ and rotational speed of thrusters $\omega_{t}$ as illustrated in Fig. \ref{schematic_depiction_usv}. Hence the mathematical model of the USV is expressed as: 
 
 \begin{equation}
    \dot{\overrightarrow{X}} = \begin{bmatrix} \dot{E}\\ \dot{N}\\ \dot{\psi}\\ \end{bmatrix} = \begin{bmatrix} u\cos{\psi} - v\sin{\psi}\\ u\sin{\psi} + v\cos{\psi}\\ r\\ \end{bmatrix},
 \end{equation}
 \noindent where 
 \begin{equation}
    \begin{bmatrix} u\\ v\\ \end{bmatrix} = \begin{bmatrix} E_{v}\sin{\psi} + N_{v}\cos{\psi}\\ E_{v}\cos{\psi} + N_{v}\sin{\psi}\\ \end{bmatrix},
 \end{equation}
 \noindent where $E_{v}$ is the velocity component of $V$ along due east and $N_{v}$ is the velocity component of $V$ along due north.

 \begin{table}[t!]
 \caption{WAM-V 20 USV specifications \cite{wamv_20}.}
 \label{Table:wamv}
 \centering
 \scriptsize
 \begin{tabular}{|c|c|c|c|}
 \hline
 \textbf{Vehicle Length} & \textbf{Vehicle Width} & \textbf{Vehicle Weight} & \textbf{Maximum Speed} \\ \hline
 6 [m] & 3 [m] & 320 [kg] & 10 [m/s] \\ \hline
 \end{tabular}
 \end{table}

 \begin{figure*}[t!]
 \centering
 \includegraphics[width=1.0\linewidth]{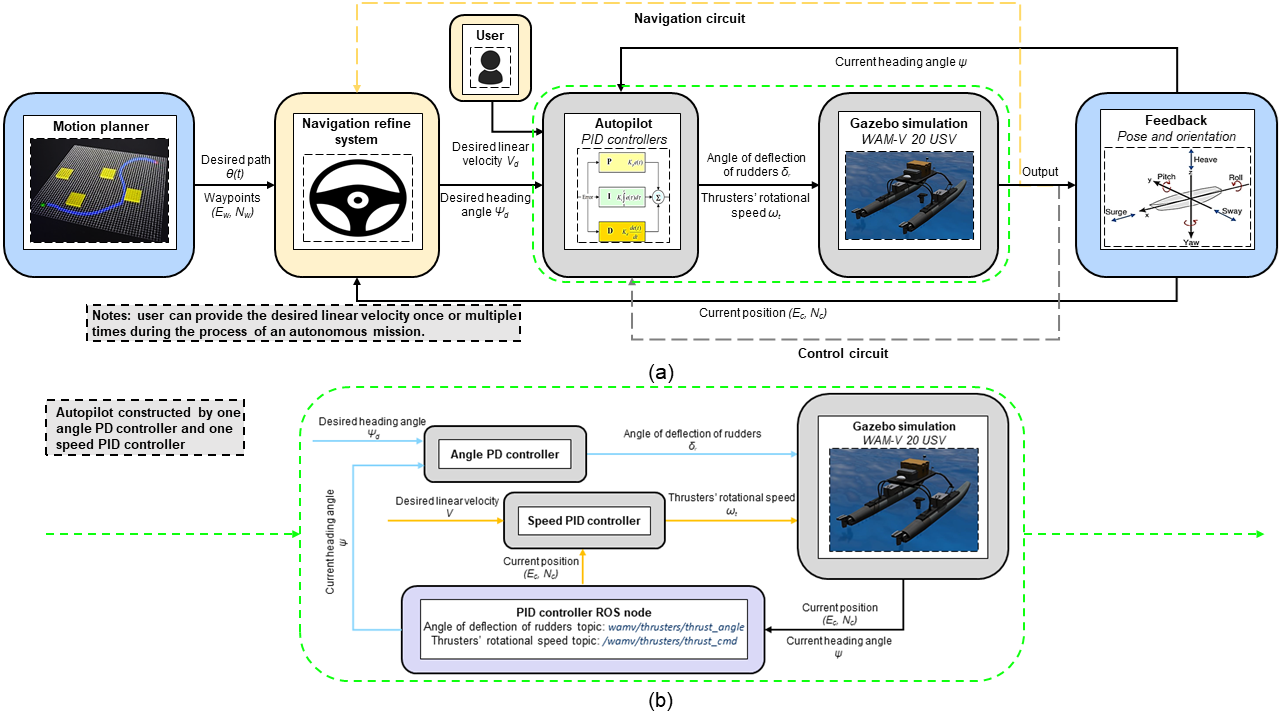}
 \centering
 \caption{(a) Overall structure of the proposed fully-autonomous USV framework (b) Detailed structure of the two controllers used in the proposed autopilot. \textcolor{black}{The operator needs only to specify the target goal point and the catamaran's linear speed before the framework starts. This framework then generates a collision-free and smooth path between the current position of the catamaran and any goal point present in the Gazebo environment and makes the catamaran follow this generated path automatically without requiring any operator interaction.}}
 \label{system_architectures} 
 \end{figure*}

\subsection{Navigation refinement system}
 The navigation refinement system can provide a timely adjustments for the USV while tracking the desired path \textcolor{black}{based upon the Light-of-sight (LOS) algorithm \cite{fossen2003line}.} The system uses the position of the next waypoint and the current position of the USV to determine the required update to the heading angle of the USV. Given the path generated by the motion planner as:
 \begin{equation}
 \theta(t) = [(E_{w_{1}}, N_{w_{1}}),...(E_{w_{n}}, N_{w_{n}})],
 \end{equation}
 %\begin{equation}
 %\theta(t_{i}) = [(E_{w_{i}}, N_{w_{i}})],
 %\end{equation}
 \noindent where $(E_{w_{1}}, N_{w_{1}})$ is the first waypoint on the desired path, $(E_{w_{i}}, N_{w_{i}})$ is the $i$th waypoint on the desired path, $(E_{w_{n}}, N_{w_{n}})$ is the last waypoint on the desired path and the path generated by the motion planner $\theta(t)$ is a function of time. Hence at a certain moment $t$, the position of the next desired waypoint $(E_{w}, N_{w})$ can be found. 
 
 The reference frame in the Gazebo virtual world is expressed as $\textbf{G} = \{X, Y\}$. As can be seen in Fig. \ref{schematic_depiction_usv}, the direction of the X axis in $\textbf{G}$ coincides with the direction of N axis in $\textbf{N}$ and the direction of the Y axis in $\textbf{G}$ coincides with the direction of E axis in $\textbf{N}$. Furthermore, the rotational angle in $\textbf{N}$ belongs to $(0,2\pi]$ and the rotational angle in $\textbf{G}$ belongs to $(-\pi,\pi]$. The rotational angles in the $\textbf{G}$ and $\textbf{N}$ reference frames need to be made uniform prior to obtaining the current position of the USV in the Gazebo virtual world. By inputting the position of the next waypoint and the current position of the USV into the navigation refinement system, the next desired heading angle of the USV can be obtained.

\subsection{Autopilot}
 To track the desired path accurately and smoothly, it is necessary to build a high-performance control mechanism to minimise the deviation between the planned path and the actual path. Based on the mechanical structure of the selected USV, two separate controllers need to be designed as: 1) an angle controller responsible for adjusting the angle of deflection of rudders (or the USV's yaw speed) and 2) a speed controller responsible for adjusting the rotational speed of thrusters (or the USV's linear speed within E-N plane). The overall structure of the proposed fully-autonomous USV navigation control system is detailed in Fig. \ref{system_architectures} (a), while the communicating and interfacing arrangement of the controllers used in the proposed autopilot is detailed in Fig. \ref{system_architectures} (b). \textcolor{black}{Proportional–integral–derivative} (PID) control is used for designing the two controllers as it has been widely adopted in previous practical USV applications \cite{liu2017fast, song2019smoothed, klinger2013controller}. Other types of controllers, such as back-stepping \cite{zhou2020adaptive, klinger2014experimental, klinger2016control} and finite-time path-following \cite{wang2019surge}, can be modified to be used as long as correct ROS messages are communicated. More details regarding the fine-tuned autopilot can be found in the open source library at: \url{https://github.com/jiaweimeng/wam-v-autopilot} 
 
 \subsubsection{Angle PD controller}
 \textcolor{black}{It is a PD controller with tuned parameters (P = 1.5 and D = 12.5). This PD controller is used to adjust the USV's rudder angle to match the desired rudder angle according to the waypoints on the desired path. Compared with the standard PID controller, we excluded the integration term as we discovered no explicit steady-state error between the current and the desired rudder angles after turning.}
 
 \textcolor{black}{To follow an arbitrary smooth path, the desired rudder angle is one of the controller inputs used to calculate the orientation error:
 \begin{equation}
  e_{\Delta\psi} = \psi - \psi_{d}
 \end{equation}
 where $\psi$ is the USV's current rudder angle, $\psi_{d}$ is the desired rudder angle and the ranges of $\psi$ and $\psi_{d}$ are $(-\pi, \pi]$.}
 
 \textcolor{black}{Based on a real-time acquired orientation error, an angle PD controller can then be constructed in the continuous-time domain:
 \begin{equation}
  \delta_{r} = k_{p}[e_{\Delta\psi_{t}}] + k_{d}[\frac{d(e_{\Delta\psi_{t}})}{dt}],
 \end{equation}
 where $k_{p}$ and $k_{d}$ are the PD gains, $\delta_{r}$ is the angle of deflection, [$e_{\Delta\psi_{t}}$] is the proportional error and [$\frac{d(e_{\Delta\psi_{t}})}{dt}$] is the differential error.} 
 
 \textcolor{black}{Due to the entire fully-autonomous USV system is built in discrete-time domain, it can then be expressed as:
 \begin{equation}
  \delta_{r} = k_{p}[e_{\Delta\psi_{i}}] + k_{d}[(e_{\Delta\psi_{i}} - e_{\Delta\psi_{i-1}})],
 \end{equation}
 where $k_{p}$ and $k_{d}$ are the PD gains, $\delta_{r}$ is the angle of deflection, [$e_{\Delta\psi_{i}}$] and [$(e_{\Delta\psi_{i}} - e_{\Delta\psi_{i-1}})$] are the corresponding proportional error and differential error in the discrete-time domain, respectively.}

 \subsubsection{Speed PID controller}
 \textcolor{black}{It is a PID controller with tuned parameters (P = 2.5, I = 0.05 and D = 1.7). This PID controller is used to adjust the thrusters' rotational speed, hence to match the actual linear velocity of the USV with the desired value according to the user's requirement.}

 \textcolor{black}{To maintain the actual velocity of the USV just at the level of the desired linear velocity or the user-specified velocity, the actual velocity of the USV is one of the controller inputs used to calculate the velocity error:
 \begin{equation}
  e_{\Delta V} = V - V_{d}
 \end{equation}
 where $V$ is the actual linear velocity of the USV, $V_{d}$ is the desired linear velocity and the ranges of them will be described in Section. \ref{ROS}.}
 
 \textcolor{black}{Nevertheless, the actual linear velocity of the USV cannot be obtained straightforwardly from the Gazebo simulation environment. Thus we need to measure it through the following equation:
 \begin{equation}
  V = \frac{\sqrt{(N_{c}-N_{p})^{2} + (E_{c}-E_{p})^{2}}}{\Delta T},
 \end{equation}
 where ($E_{c}, N_{c}$) and ($E_{p}, N_{p}$) are the current position and the previous position of the USV obtained straightforwardly from the Gazebo simulation environment between one system interval period $\Delta T$, respectively.}
 
 \textcolor{black}{Based on a real-time acquired velocity error, a speed PID controller can then be constructed in the continuous-time domain:
 \begin{equation}
 \omega_{t} = k_{p}[e_{\Delta V_{t}}] + k_{i}[\int_{0}^{t_{c}}(e_{\Delta V_{t}})\,dt] + k_{d}[\frac{d(e_{\Delta V_{t}})}{dt}],
 \end{equation}
 where $k_{p}$, $k_{i}$ and $k_{d}$ are the PID gains, $\omega_{t}$ is the thrusters' rotational speed of the USV, [$e_{\Delta V_{t}}$] is the proportional error, [$\int_{0}^{T_{c}}(e_{\Delta V_{t}})\,dt$] is the integral error, [$\frac{d(e_{\Delta V_{t}})}{dt}$] is the differential error and $t_{c}$ is the present moment.} 
 
 \textcolor{black}{Due to the entire fully-autonomous USV system is built based on the discrete-time domain, it can then be expressed as:
 \begin{equation}
 \omega_{t} = k_{p}[e_{\Delta V_{i}}] + k_{i}[\sum_{n=0}^{T_{i}} e_{\Delta V_{t}}] + k_{d}[(e_{\Delta V_{i}} - e_{\Delta V_{i-1}})],
 \end{equation}
 where $k_{p}$, $k_{i}$ and $k_{d}$ are the PID gains, $\omega_{t}$ is the thrusters' rotational speed of the USV, [$e_{\Delta\psi_{i}}$], [$\sum_{n=0}^{T_{i}} e_{\Delta V_{t}}$] and [$(e_{\Delta\psi_{i}} - e_{\Delta\psi_{i-1}})$] are the corresponding proportional error, integral error and differential error in the discrete-time domain, respectively.}

\section{Simulations and discussions}
 This section demonstrates the performance of the proposed motion planning algorithm on the basis of comparisons against three simulation benchmarks.

 \begin{table}[t!]
 \caption{Specification of the parameters used in the motion planning algorithms.}
 \label{Table:2}
 \centering
 \footnotesize
 \begin{tabular}{|c|ccccc|c|c|}
 \hline
 \makecell{\textbf{Map [pixel]}} & \multicolumn{5}{c|}{\textbf{GP-based Motion Planning}} & \multicolumn{1}{c|}{\textbf{A*}} & \multicolumn{1}{c|}{\textbf{RRT*}} \\
 \cline{2-8}
 {} & \makecell{$\epsilon$} & \makecell{$\sigma_{obs}$} & \makecell{$\sigma_{e}$} & \makecell{$T_{max}$} & \makecell{$N$} & \makecell{$l$} & \makecell{$l$} \\
 \hline
  500x500 & 20 & 0.05 & 0.005 & 2.0 & 5 & 10.0 & 10.0 \\ \hline
  1000x1000 & 20 & 0.05 & 0.005 & 4.0 & 10 & 10.0 & 10.0 \\ \hline
  2000x2000 & 20 & 0.05 & 0.005 & 8.0 & 20 & 10.0 & 10.0 \\ \hline
     \multicolumn{8}{p{240pt}}{\textbf{{Notes:}} {\textcolor{black}{These parameters are empirically determined as values provide a good trade-off between collision avoidance and energy consumption reduction.}}}
 \end{tabular}
 \end{table}

 \begin{table}[t!]
 \caption{Specification of the used hardware platform.}
 \label{Table:3}
 \centering
 %\small
 \footnotesize
 \begin{tabular}{|c|c|c|}
 \hline
 \textbf{Name of the Device} & \textbf{Description} & \textbf{Quantity} \\ \hline
 {Processor} & 2.6-GHz Intel Core i7-6700HQ & 8 \\ \hline
 {RAM} & 8 GB & 1 \\ \hline
 \end{tabular}
 \end{table}

 \subsection{Simulation details}
 Three simulation benchmarks have been conducted to evaluate the proposed MC-GPMP2*. First, the incremental optimisation process of the proposed method was subjected to qualitative tests. Then the proposed method was quantitatively compared with other state-of-the-art motion planning algorithms including GPMP2 \cite{dong2016motion}, A* \textcolor{black}{(or A star)} \cite{hart1968formal}, RRT* (or rapidly-exploring random tree star) \cite{lavalle1998rapidly} and AFM (or anisotropic fast marching method) \cite{lin2003enhancement} in different environments both with and without environment characteristics (ocean currents). In all the simulations, GP-based methods were always initialised with a constant-velocity straight-line trajectory. Table \ref{Table:2} details the specifications of the parameters used in the motion planning algorithms. The specific parameters of GP-based motion planning, A* and RRT* in the following simulations in various resolutions are clarified. In Table \ref{Table:2}, $\epsilon$ indicates the safety distance [pixel], $\sigma_{obs}$ indicates the obstacle cost weight, $\sigma_{e}$ indicates the energy cost weight, $T_{max}$ indicates the total sampling time [s], $N$ indicates the low-resolution region number in Algorithm \ref{alg:2} and $l$ indicates the step size [pixel]. In the following simulations, one pixel in the map equals one meter in the corresponding motion planning problem. Table \ref{Table:3} is a specification of the hardware platform used. 

 \subsection{System dynamics model}
 Applying a constant-velocity motion model minimises acceleration along the trajectory, thus reducing energy consumption and increasing the smoothness of the generated path. The system dynamics of the robot platform is represented with the double integrator linear system with additional white noise on acceleration. The trajectory is then generated by (\ref{LTV_SDE}) with the following specific parameters:
 
 \begin{figure*}[t!]
 \centering
 \includegraphics[width=1.0\linewidth]{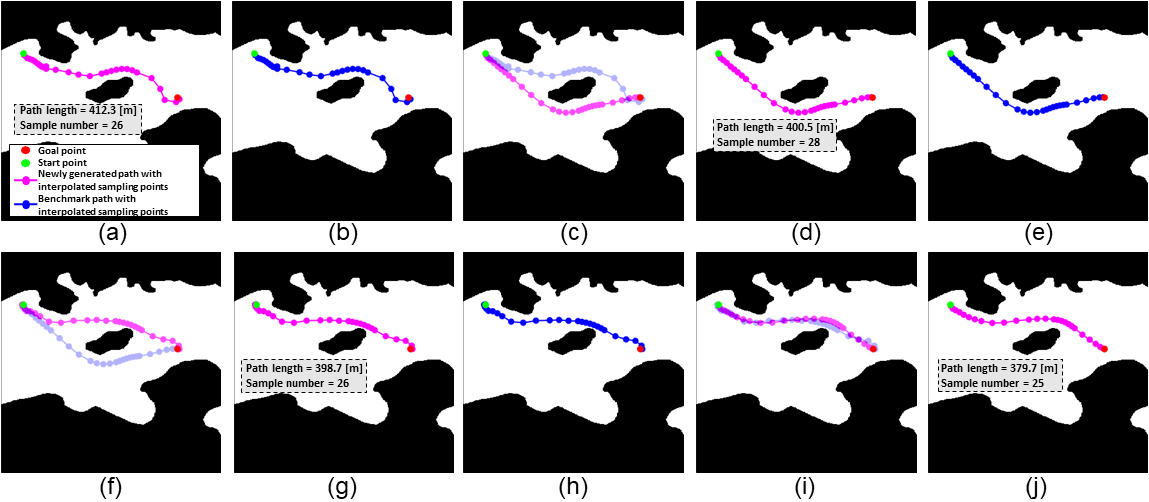}
 \centering
 \caption{A demonstration of the incremental optimisation process of MC-GPMP2* in replanning problems: (a) a new path is generated, (b) this newly generated path turns into a benchmark path, (c) this benchmark path is compared with another newly generated path and the latter will be accepted if 1) it is shorter than the former and 2) it does not collide with any obstacle and (d) the newly generated path is accepted. Similar to (a) - (d), (d) - (g) and (g) - (j) repeat the process to achieve incremental optimisation in replanning problems. Overall, the path length generated by MC-GPMP2* decreases from 412.3 [m] to 379.7 [m] within 5 replanning iterations.}
 \label{optimisation_process} 
 \end{figure*}
 
 \begin{equation}
    A = \begin{bmatrix}0 & I\\ 0 & 0\end{bmatrix}, x(t) = \begin{bmatrix}r(t)\\v(t)\end{bmatrix}, F(t) = \begin{bmatrix}0\\I\end{bmatrix}, u(t) = 0, 
 \end{equation}
 where $r = (x, y)^T$ is the position vector, $v = (v_x, v_y)$ is the velocity vector and given $\Delta t_{i} = t_{i+1} - t_{i}$,
  \begin{equation}
    \Phi(t,s) = \begin{bmatrix}I & (t-s)I\\ 0 & I\end{bmatrix}, Q_{i,i+1} = \begin{bmatrix} \frac{1}{3}\Delta t_{i}^{3}Q_{C} & \frac{1}{2}\Delta t_{i}^{2}Q_{C} \\ \frac{1}{2}\Delta t_{i}^{2}Q_{C} & \Delta t_{i}Q_{C} \end{bmatrix},
 \end{equation}
 This prior is centred around a zero-acceleration trajectory \textcolor{black}{(or a straight-line segment)} \cite{dong2016motion}. \textcolor{black}{During the optimisation process, the cost function can make the trajectory deviate from the straight-line segment to construct an optimised trajectory.}
 
\subsection{Incremental optimisation process of the proposed method}
 
 In this subsection, we demonstrate the incremental optimisation process of the proposed GPMP2* when trajectory replanning is taking place in a coastal region. We explicitly reveal how the Monte-Carlo sampling can adaptively vary the number of sampling points to generate an optimised trajectory. As shown in Fig. \ref{optimisation_process}, by having 5 support states, a new path with 26 sampling points is generated as shown in Fig. \ref{optimisation_process} (a) with the path length being 412.3 [m]. The number of sampling points between each support state are 7, 4, 3, 8 and 4, respectively.
 By using this path as a benchmark (Fig. \ref{optimisation_process} (b)), a new path with 28 sampling points is generated as shown in Fig. \ref{optimisation_process} (c) with the length being 400.5 [m] and the sampling points between each support state being 7, 4, 3, 10 and 4, respectively. A comparison between the benchmark path and this new path is then conducted. The new path will be accepted if 1) it is shorter than the benchmark path and 2) it does not intersect with any obstacle. Such iterative comparisons will continue until no more new paths are generated and an optimal trajectory can then be selected, which in this case is that shown in Fig. \ref{optimisation_process} (j). 
 
 To summarise, GP-based motion planning generates a trajectory from a stochastic process, where the pattern of the trajectory is determined by the sampling points. Within the conventional GP-based motion planning, such as GPMP2, although an option to adjust the number of sample points is provided, there is a lack of strategy to achieve the optimal number of sample points, forcing most GP-based motion planning algorithms to require manual tuning of the number of sample points. Monte Carlo stochasticity can be added to GP-based motion planning algorithms to achieve an adaptive tuning process by doing the following strategy: within a region with a small number of support states, more states can be interpolated based upon the number of obstacles, i.e. a larger number of sample points would need to be interpolated to deal with a number of densely packed obstacles while reducing the number of points for less densely packed obstacles. By following such a strategy, sampling points can be adjusted and interpolated more effectively and efficiently. 
 
 \begin{figure*}[t!]
 \centering
 \includegraphics[width=1.0\linewidth]{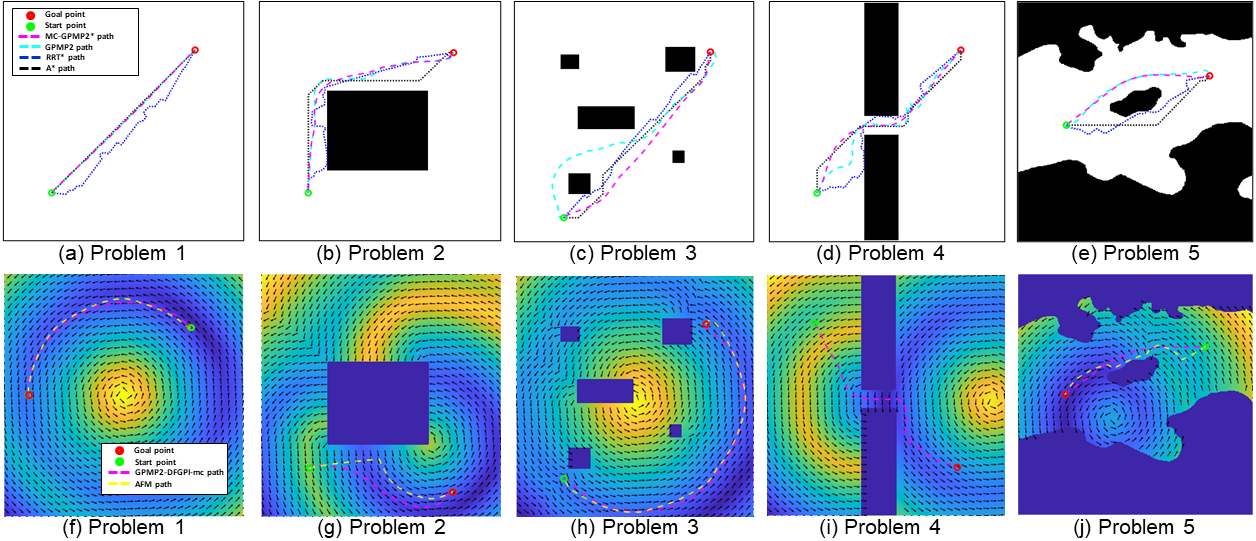}
 \centering
 \caption{Comparisons of the paths generated by various motion planning algorithms in different scenarios with and without environment characteristics from 500 * 500 pixel maps: (a) and (f) demonstrate non-obstacle scenario (problem 1), (b) and (g) demonstrate single-obstacle scenario (problem 2), (c) and (h) demonstrate multi-obstacle scenario (problem 3), (d) and (i) demonstrate narrow-passage scenario (problem 4), (e) and (j) demonstrate coastal scenario (problem 5). Furthermore, (a) - (e) have no ocean currents but (f) - (j) have ocean currents.}
 \label{simulations_w&wo_currents} 
 \end{figure*}
 
 \begin{table*}[t!]
 \caption{A comparison of MC-GPMP2*, GPMP2, A* and RRT* on average execution time ($T$) and path length ($L$) in 15 path planning problems without ocean currents. The improvement on average execution time ($T_{I}$) with the Monte Carlo stochasticity was also measured in each path planning problem. \textcolor{black}{The experiment on each path planning problem was tested 5 times to calculate the average value.}}
 \label{Table:4}
 \centering
 \footnotesize
 %\rotatebox{90}{
 \begin{tabular}{|m{1.65cm}<{\centering}|m{1.4cm}<{\centering}|m{1.15cm}<{\centering}m{1.15cm}<{\centering}m{1.15cm}<{\centering}|m{1.15cm}<{\centering}m{1.15cm}<{\centering}|m{1.15cm}<{\centering}m{1.15cm}<{\centering}|m{1.15cm}<{\centering}m{1.15cm}<{\centering}|}
 \hline
 \textbf{Map [pixel]} & \textbf{Problem} & \multicolumn{3}{c|}{\cellcolor{green!25}\textbf{MC-GPMP2*}} & \multicolumn{2}{c|}{\textbf{GPMP2}} &  \multicolumn{2}{c|}{\textbf{A*}} & \multicolumn{2}{c|}{\textbf{RRT*}} \\
 \cline{3-11}
 {} & {} & {$T$ [ms]} &{$L$ [m]} & {$T_{I}$ [ms]} & {$T$ [ms]} & {$L$ [m]} & {$T$ [ms]} & {$L$ [m]} & {$T$ [ms]} & {$L$ [m]} \\
 \hline
  &  1  & \cellcolor{cyan!25}202.1 & 500.8 & \cellcolor{yellow!25}58.0 & 283.3 & 500.8 & 1847.1 & \cellcolor{cyan!25}494.9 & 3261.4 & 562.1 \\ 
  &  2  & \cellcolor{cyan!25}154.3 & \cellcolor{cyan!25}607.2 & \cellcolor{yellow!25}39.9 & 208.5 & 618.5 & 21804.8 & 617.9 & 4185.8 & 656.5 \\ 
  500x500 &  3  & \cellcolor{cyan!25}175.7 & \cellcolor{cyan!25}521.9 & \cellcolor{yellow!25}49.6 & 236.5 & 532.7 & 15204.6 & 529.8 & 3723.1 & 577.5 \\ 
  &  4  & \cellcolor{cyan!25}208.8 & 480.2 & \cellcolor{yellow!25}67.8 & 292.3 & 491.2 & 13430.5 & \cellcolor{cyan!25}476.9 & 3633.3 & 610.2 \\ 
  &  5  & \cellcolor{cyan!25}187.5 & \cellcolor{cyan!25}234.6 & \cellcolor{yellow!25}17.3 & 224.5 & 245.1 & 6834.5 & 283.1 & 2595.8 & 322.5 \\ \hline
  &  1  & \cellcolor{cyan!25}214.6 & 992.1 & \cellcolor{yellow!25}69.2 & 306.3 & 992.1 & 4307.4 & \cellcolor{cyan!25}989.9 & 7343.5 & 1126.4 \\ 
  &  2  & \cellcolor{cyan!25}207.5 & \cellcolor{cyan!25}1245.5 & \cellcolor{yellow!25}49.3 & 277.9 & 1267.1 & - & - & 10388.7 & 1360.2 \\ 
  1000x1000 &  3  & \cellcolor{cyan!25}214.9 & \cellcolor{cyan!25}1104.7 & \cellcolor{yellow!25}57.4 & 286.3 & 1119.3 & - & - & 6736.4 & 1137.1 \\ 
  &  4  & \cellcolor{cyan!25}230.5 & \cellcolor{cyan!25}1107.2 & \cellcolor{yellow!25}80.5 & 339.2 & 1120.1 & - & - & 8366.2 & 1406.3 \\ 
  & 5  & \cellcolor{cyan!25}233.8 & \cellcolor{cyan!25}546.8 & \cellcolor{yellow!25}31.1 & 287.3 & 562.7 & 17434.5 & 549.7 & 4823.3 & 562.6 \\ \hline
  & 1  & \cellcolor{cyan!25}363.1 & 1981.5 & \cellcolor{yellow!25}82.6 & 471.1 & 1981.5 & 5748.3 & \cellcolor{cyan!25}1979.9 & 15216.2 & 2275.5 \\ 
  & 2  & \cellcolor{cyan!25}340.9 & \cellcolor{cyan!25}2432.9 & \cellcolor{yellow!25}61.3 & 427.2 & 2499.6 & - & - & 22966.2 & 2665.7 \\ 
  2000x2000 & 3 & \cellcolor{cyan!25}369.8 & \cellcolor{cyan!25}2201.5 & \cellcolor{yellow!25}79.6 & 463.9 & 2222.7 & - & - & 19636.8 & 2273.7 \\ 
  &  4  & \cellcolor{cyan!25}358.5 & \cellcolor{cyan!25}2028.8 & \cellcolor{yellow!25}105.8 & 497.9 & 2045.5 & - & - & 18707.3 & 2383.2 \\ 
  &  5  & \cellcolor{cyan!25}406.6 & \cellcolor{cyan!25}1178.7 & \cellcolor{yellow!25}51.6 & 491.5 & 1195.3 & - & - & 11049.9 & 1452.2 \\ \hline 
 \multicolumn{11}{p{500pt}}{\textbf{{Notes:}} {"-" means the motion planning algorithm is not applicable in this map as its execution time is more than 30 [s], which is meaningless in practical situations. The proposed method is marked in light green. The shortest execution time ($T$) and the shortest path length ($L$) in each problem are marked in light blue. Meanwhile, the improvement on average execution time ($T_{I}$) with Monte-Carlo stochasticity in each problem is marked in light yellow. Without Monte-Carlo stochasticity, MC-GPMP2* uses a traversal algorithm to estimate the obstacle space. In this benchmark, all the motion planning algorithms only run once, which means the replanning processes of them are excluded. For instance, the re-wiring process of the tree branches of RRT* will be terminated once a feasible path has been found.
 }}
 \end{tabular}
 %}
 \end{table*}
 
 \begin{table*}[t!]
 \caption{A comparison of MC-GPMP2* and AFM on average energy consumption rate ($P$), execution time ($T$) and path length ($L$) in 15 path planning problems with ocean currents. The improvement on average execution time ($T_{I}$) with the Monte Carlo stochasticity was also measured in each path planning problem. \textcolor{black}{The experiment on each path planning problem was tested 5 times to calculate the average value.}}
 \label{Table:5}
 \centering
 \footnotesize
 %\rotatebox{90}{
 \begin{tabular}{|m{2.0cm}<{\centering}|m{1.7cm}<{\centering}|m{1.5cm}<{\centering}m{1.5cm}<{\centering}m{1.5cm}<{\centering}m{1.5cm}<{\centering}|m{1.5cm}<{\centering}m{1.5cm}<{\centering}m{1.5cm}<{\centering}|}
 \hline
 \textbf{Map [pixel]} & \textbf{Problem} & \multicolumn{4}{c|}{\cellcolor{green!25}\textbf{MC-GPMP2*}} &  \multicolumn{3}{c|}{\textbf{AFM}} \\
 \cline{3-9}
 {} & {} & {$P$ [\%]} & {$T$ [ms]} &{$L$ [m]} & {$T_{I}$ [ms]} & {$P$ [\%]} &{$T$ [ms]} & {$L$ [m]} \\
 \hline
  &  1  & 11.8 & \cellcolor{cyan!25}684.2 & \cellcolor{cyan!25}327.2 & \cellcolor{yellow!25}59.4 & \cellcolor{cyan!25}10.5 & 909.6 & 344.6 \\ 
  &  2  & 4.2 & \cellcolor{cyan!25}608.0 & \cellcolor{cyan!25}154.5 & \cellcolor{yellow!25}47.3 & \cellcolor{cyan!25}2.1 & 943.6 & 167.4 \\ 
  500x500 &  3  & 15.8 & \cellcolor{cyan!25}682.5 & \cellcolor{cyan!25}463.6 & \cellcolor{yellow!25}77.6 &  \cellcolor{cyan!25}4.9 & 815.7 & 505.3 \\ 
  &  4  & \cellcolor{cyan!25}2.0 & \cellcolor{cyan!25}646.8 & \cellcolor{cyan!25}168.6 & \cellcolor{yellow!25}115.8 & - & - & - \\ 
  &  5  & 5.3 & \cellcolor{cyan!25}609.2 & \cellcolor{cyan!25}236.7 & \cellcolor{yellow!25}51.2 & \cellcolor{cyan!25}2.7 & 715.6 & 258.8 \\ \hline
  &  1  & 12.1 & \cellcolor{cyan!25}2401.5 & \cellcolor{cyan!25}658.9 & \cellcolor{yellow!25}103.1 & \cellcolor{cyan!25}10.5 & 2559.8 & 684.8 \\ 
  &  2  & 4.2 & \cellcolor{cyan!25}2195.1 & \cellcolor{cyan!25}309.1 & \cellcolor{yellow!25}107.4 & \cellcolor{cyan!25}2.0 & 2306.5 & 336.5 \\ 
  1000x1000 &  3  & 14.1 & \cellcolor{cyan!25}2514.7 & \cellcolor{cyan!25}946.9 & \cellcolor{yellow!25}143.6 & \cellcolor{cyan!25}4.9 & 2731.2 & 1014.6 \\ 
  &  4  & \cellcolor{cyan!25}2.0 & \cellcolor{cyan!25}2181.4 & \cellcolor{cyan!25}355.6 & \cellcolor{yellow!25}197.0 & - & - & - \\ 
  & 5  & 5.0 & \cellcolor{cyan!25}2112.3 & \cellcolor{cyan!25}480.8 & \cellcolor{yellow!25}67.4 & \cellcolor{cyan!25}2.7 & 2265.3 & 517.2 \\ \hline
  & 1  & 11.6 & \cellcolor{cyan!25}10821.3 & \cellcolor{cyan!25}1306.8 & \cellcolor{yellow!25}166.9 & \cellcolor{cyan!25}10.3 & 11193.4 & 1364.3 \\ 
  & 2  & 2.2 & \cellcolor{cyan!25}9263.9 & \cellcolor{cyan!25}407.9 & \cellcolor{yellow!25}208.2 & \cellcolor{cyan!25}1.0 & 9536.5 & 442.3 \\ 
  2000x2000 & 3 & 14.9 & \cellcolor{cyan!25}11504.6 & \cellcolor{cyan!25}1821.8 & \cellcolor{yellow!25}222.8 & \cellcolor{cyan!25}4.9 & 11576.5 & 2027.8 \\ 
  &  4  & \cellcolor{cyan!25}2.1 & \cellcolor{cyan!25}9724.3 & \cellcolor{cyan!25}730.3 & \cellcolor{yellow!25}252.5 & - & - & - \\
  &  5  & 5.3 & \cellcolor{cyan!25}9006.2 & \cellcolor{cyan!25}954.9 & \cellcolor{yellow!25}155.4 & \cellcolor{cyan!25}2.7 & 9245.6 & 1034.0 \\ \hline
 \multicolumn{9}{p{500pt}}{\textbf{{Notes:}} {"-" means the motion planning algorithm is not applicable in this map as its execution time is more than 30 [s], which is meaningless in practical situations. The energy consumption rate ($P$) caused by ocean currents is computed based upon the metric proposed in AFM \cite{song2017multi} as explained in (\ref{environment_compensation_function}). The proposed method is marked in light green. The shortest execution time ($T$) and the shortest path length ($L$) in each problem are marked in light blue. Meanwhile, the improvement on average execution time ($T_{I}$) with Monte-Carlo stochasticity in each problem is marked in light yellow. Without Monte-Carlo stochasticity, MC-GPMP2* uses traversal algorithm to estimate obstacle space. The replanning process of MC-GPMP2* is excluded in this benchmark.}}
 \end{tabular}
 %}
 \end{table*}
 
\subsection{Benchmark without environment characteristics}
\label{benchmark_without_oc}
 In this subsection, we conduct a comparative study showing the improvement of MC-GPMP2* against the mainstream motion planning algorithms such as GPMP2, A* and RRT*. Various simulation environments are adopted including: 1) a no-obstacle environment, 2) a single-obstacle environment, 3) a multi-obstacle environment, 4) a narrow-passage environment and 5) a coastal environment without any environment characteristics. Note that within the MC-GPMP2*, a relatively large number of sampling points is used to guarantee the generation of optimised trajectories.
 
 The simulation results are shown in Fig. \ref{simulations_w&wo_currents} (a) - (e). Note that only the results from the 500 * 500 pixel maps are illustrated as different resolutions mainly affects the computation time rather than the generated trajectories. A quantitative assessment of different algorithms is shown in Table \ref{Table:4}, where main evaluation metrics such as execution time and path length are compared.
 
 From Fig. \ref{simulations_w&wo_currents} (a) - (e), MC-GPMP2* illustrates a distinct advantage regarding the average path length and path smoothness compared with GPMP2, A* and RRT*. In no-obstacle environments (problem 1), MC-GPMP2*, GPMP2 and A* each generate a straight-line path that connects the start point and goal point simply by the shortest distance. However, RRT* generates a winding path with the longest path length and lowest path smoothness. In single-obstacle environments (problem 2), the paths generated by MC-GPMP2* and GPMP2 are of relatively short length and relatively high smoothness. Compared with the GPMP2 path, the MC-GPMP2* path shows further improvement in both the path length and smoothness. This is a benefit of the proposed interpolation strategy. On the other hand, both the paths generated by A* and RRT* are as smooth and might not be smooth enough to satisfy the system dynamics model of the USV. In multi-obstacle environments (problem 3), MC-GPMP2* and GPMP2 paths demonstrate better path smoothness based upon a comparison with the A* and RRT* paths. However, the GPMP2 path tends to avoid the first obstacle sweeping out around the left hand side, leading to a significant increase in the path length. With the proposed interpolation strategy, MC-GPMP2* generates an option that would avoid the first obstacle from the bottom side and this results in a decrease on the path length. Similar to multi-obstacle environments (problem 3), MC-GPMP2* produces a path option which presents a further improvement on length and smoothness compared with the GPMP2 path in narrow-passage environments (problem 4). This is because most of the sampling points of MC-GPMP2* were sampled around the narrow passage to improve the option for success of the mission and shorten the length of the path apart from the narrow passage itself. In coastal environments, MC-GPMP2* demonstrates the highest path smoothness and the best obstacle avoidance performance as would be expected.
 
 From Table \ref{Table:4}, MC-GPMP2* demonstates an obvious benefit on average execution time and path length over GPMP2, A* and RRT* in maps across a range of resolutions. In most of the large-scale motion planning problems with 1000 * 1000 pixel and 2000 * 2000 pixel maps, A* failed to deliver a feasible solution. This is because the motion planning strategy of A* led to a significant increase in complexity in large-scale motion planning problems. Although RRT* could consistently deliver a feasible solution in all the motion planning problems, the average execution time, path length and path smoothness were not satisfactory as the randomness of its sampling points is too high in the configuration space. Compared with GPMP2*, the interpolation strategy of MC-GPMP2* led to a notable improvement in average execution time and path length simultaneously.
 
 To summarise, MC-GPMP2* can generate a path within the shortest execution time, with highest smoothness and near-optimal path length in almost all the cases and achieve better performance with respect to obstacle avoidance compared with other mainstream motion planning algorithms including GPMP2, A* and RRT*. 

\subsection{Benchmark with environment characteristic}
 In this subsection, we conduct another comparative study showing the improvement of MC-GPMP2* over AFM in the same simulation environments with a supplementary environment characteristic resulting from an ocean current field. The ocean current field is generated by the energy consumption metric proposed in AFM \cite{song2017multi}. 
 
 Simulation results related to this benchmark are illustrated in Fig. \ref{simulations_w&wo_currents} (f) - (j). Similar to the previous benchmark, only the results from the 500 * 500 pixel maps are shown. The quantitative assessment of MC-GPMP2* and AFM is shown in Table \ref{Table:5}, where main evaluation metrics such as energy consumption rate, execution time and path length are compared.
 
 From Fig. \ref{simulations_w&wo_currents} (f) - (j), MC-GPMP2* has an obvious advantage regarding the average execution time and path length compared with AFM. Comparatively, AFM has a considerable advantage regarding its average energy consumption rate as it continuously tracks the ocean currents. Nevertheless, this could lead to AFM falling into a local minimum when an obstacle is blocking the continuous ocean currents, such as the motion planning problems in narrow-passage environments (problem 4). MC-GPMP2* generates a path under the interaction of two different fields, namely the signed distance field and the energy consumption field. To be more precise, the signed distance field and the energy consumption field can be obtained by inputting the map in the signed distance function in (\ref{workspace_cost_function}) and the metric that can measure the energy consumption rate at each pixel in (\ref{environment_compensation_function}), respectively. Moreover, the energy consumption field can prevent the occurrence of local minima when avoiding obstacles in the signed distance field. In other words, once MC-GPMP2* has fallen into a local minimum in the signed distance field, the energy consumption field would take it out of that local minimum.
 
 From Table \ref{Table:5}, the proposed method demonstrates a notable advantage on average execution time and path length over another mainstream method (AFM) in different-resolution maps. For both methods, the energy consumption field is computed based upon the energy consumption metric stated in (\ref{environment_compensation_function}). The energy consumption field is more likely to be historical data recorded by relevant meteorological institutions. As a result, the computation time for generating the simulated energy consumption field can be saved when applying the proposed method in practical cases. This would lead to a remarkable reduction in the execution time as the proportion of the time cost on generating the energy consumption field exceeds 96 [\%] in the 2000 * 2000 pixel maps.
 
 To summarise, MC-GPMP2* can consider various environment characteristics during the motion planning process. The path generated by MC-GPMP2* would fit these characteristics as much as possible. Compared with other mainstream motion planning algorithms such as AFM, when the path planning has to adjust for the influence of ocean currents, MC-GPMP2 can generate a path with the shortest execution time, highest smoothness, near-optimal path length and a better performance on obstacle avoidance. 

 In both the benchmark tests with and without environment characteristics: the Monte-Carlo sampling algorithm can converge much earlier than the traversal algorithm, thereby reducing the time cost of a motion planner with sample points. In these benchmark tests, we only demonstrate the improvement of average execution time in 2D motion planning problems. But in high-dimensional motion planning problems, such as the motion planning problems for multiple degrees of freedom robotic arms, the Monte-Carlo sampling holds the potential to reduce a significant time cost since its convergence rate is independent of the dimension of the configuration space. Hence it solves the problem of dimensional explosion in GP-based motion planning algorithms to some extent.

 \begin{figure}[t!]
 \centering
 \includegraphics[width=1.0\linewidth]{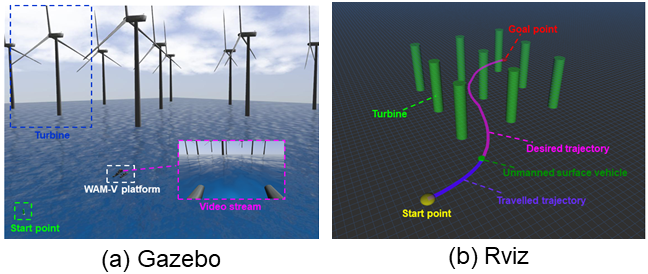}
 \centering
 \caption{ROS simulation environment: (a) demonstrates the Gazebo virtual world, where the green dash line block represents the start point, the white dash line block represents the selected platform, the blue dash line block represents the wind turbine (obstacle) and the purple dash line block represents the captured video information from the camera mounted at the front end of the platform and (b) demonstrates the corresponding motion planning problem solved by MC-GPMP2* in Rviz, where the start point is represented in yellow, the goal point is represented in red, the obstacles are represented in green, the desired path is represented in purple and the travelled path is presented in dark blue. \textcolor{black}{In the Gazebo virtual world, wind and wave fields can be adjusted by changing the corresponding parameters to create a realistic simulation environment.}}
 \label{simulation_gazebo_rviz} 
 \end{figure}

\section{Implementation in ROS}
\label{ROS}
 This section demonstrates the performance of the proposed autonomous navigation system for WAM-V 20 USVs. Two different motion planning algorithms, i.e. RRT* and the proposed MC-GPMP2*, are implemented and compared. An offshore wind farm inspection mission is simulated in ROS to show the practicability of the proposed work.
 
 \subsection{Simulation details}
 The detailed information of the environment used in the ROS simulation is detailed in Fig. \ref{simulation_gazebo_rviz}, where (a) shows the offshore wind farm in Gazebo with the inclusion of a series of physical properties such as sunlight, wind, ocean currents, gravity and buoyancy, (b) provides a simulation overview of the configuration space of the corresponding motion planning problem in Rviz. A green buoy and a red buoy are placed inside the simulation environment to indicate the start point and the goal point for the route proposed for the WAM-V 20 USV to navigate. The platform was equipped with a camera to better observe the surrounding environment and record videos. The footage from the camera was streamed to and displayed on the Rviz interface through the WAM-V Camera node (/wam-v/sensors/cameras/front-camera/image-raw). \textcolor{black}{The virtual onboard camera gives this work the potential to combine with previous research done by our research group on using onboard cameras for object detection and segmentation in maritime environments \cite{yao2021shorelinenet, chen2021wodis}.}
 
 \begin{figure}[t!]
 \centering
 \includegraphics[width=1.0\linewidth]{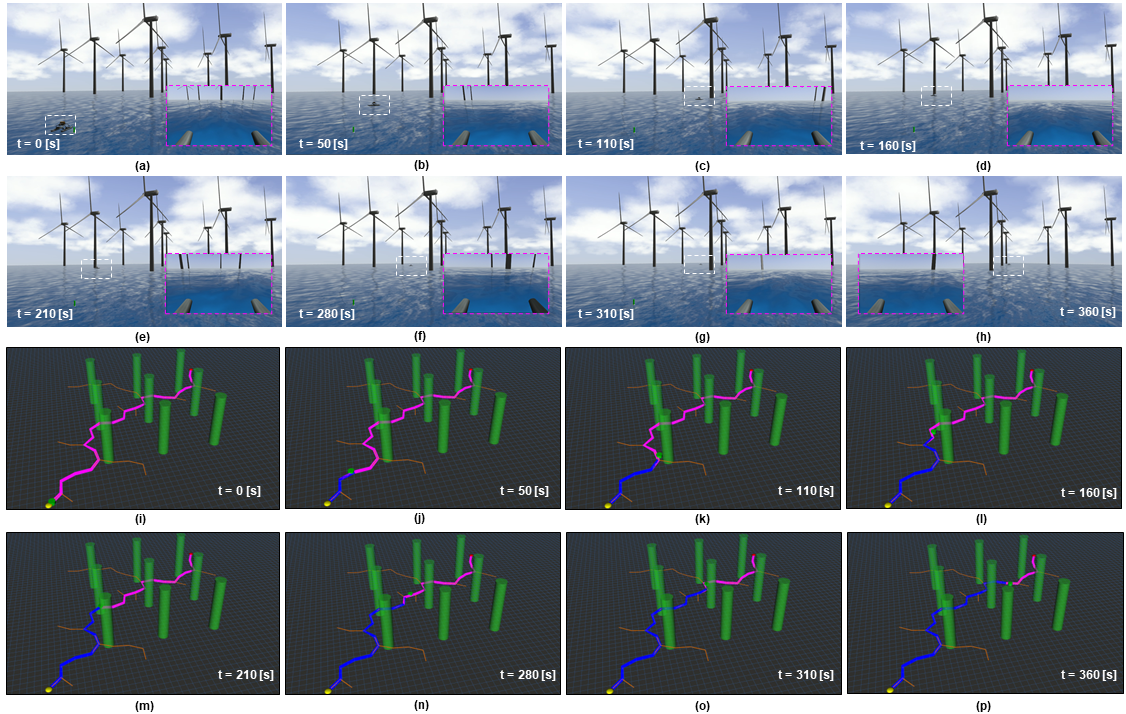}
 \centering
 \caption{The storyboards of the inspection mission in an offshore wind power generation scenario based on a path generated by RRT*: From (a) to (h), the images demonstrate the location of the platform and the video stream from the camera mounted at the front end of the platform when the time equals 0 [s], 50 [s], 110 [s], 160 [s], 210 [s], 280 [s], 310 [s] and 360 [s], respectively. From (i) to (h), the images demonstrate the corresponding motion planning problem in Rviz when the time equals 0 [s], 50 [s], 110 [s], 160 [s], 210 [s], 280 [s], 310 [s] and 360 [s], respectively.}
 \label{rrt_experiments} 
 \end{figure}
 
 \begin{figure}[t!]
 \centering
 \includegraphics[width=1.0\linewidth]{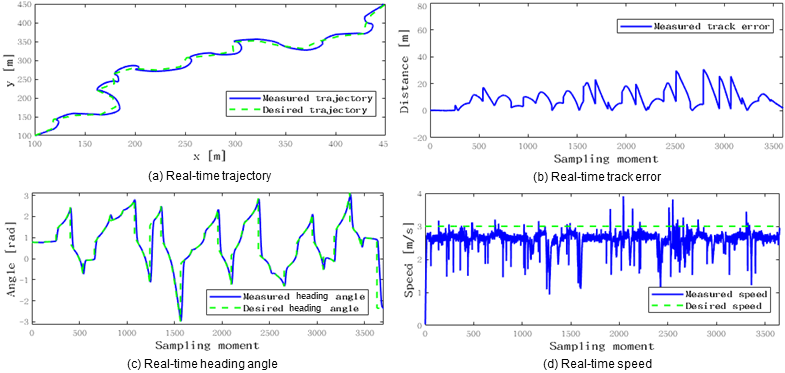}
 \centering
 \caption{Performance analysis of the inspection mission in an offshore wind power generation scenario based on a path generated by RRT*: (a) compares the desired and measured trajectories, (b) demonstrates the real-time track error, (c) compares the desired and measured heading angles and (d) compares the desired and measured speeds.}
 \label{measurements_of_rrt_path} 
 \end{figure}
 
 During the inspection mission in Gazebo, the USV transited through the wind turbine area to drive away any fish boats entering this area to reduce risk of collision and damage to the wind turbines. Figs. \ref{rrt_experiments} and \ref{gpmp2_experiments} demonstrate the storyboards of the inspection mission from both the first-person and third-person perspectives in the Gazebo as well as the motion planning problem solved by the corresponding motion planning algorithms in Rviz.
 
 In the ROS simulation, the inspection mission is designed based on the following steps: 
 
 \begin{figure}[t!]
 \centering
 \includegraphics[width=1.0\linewidth]{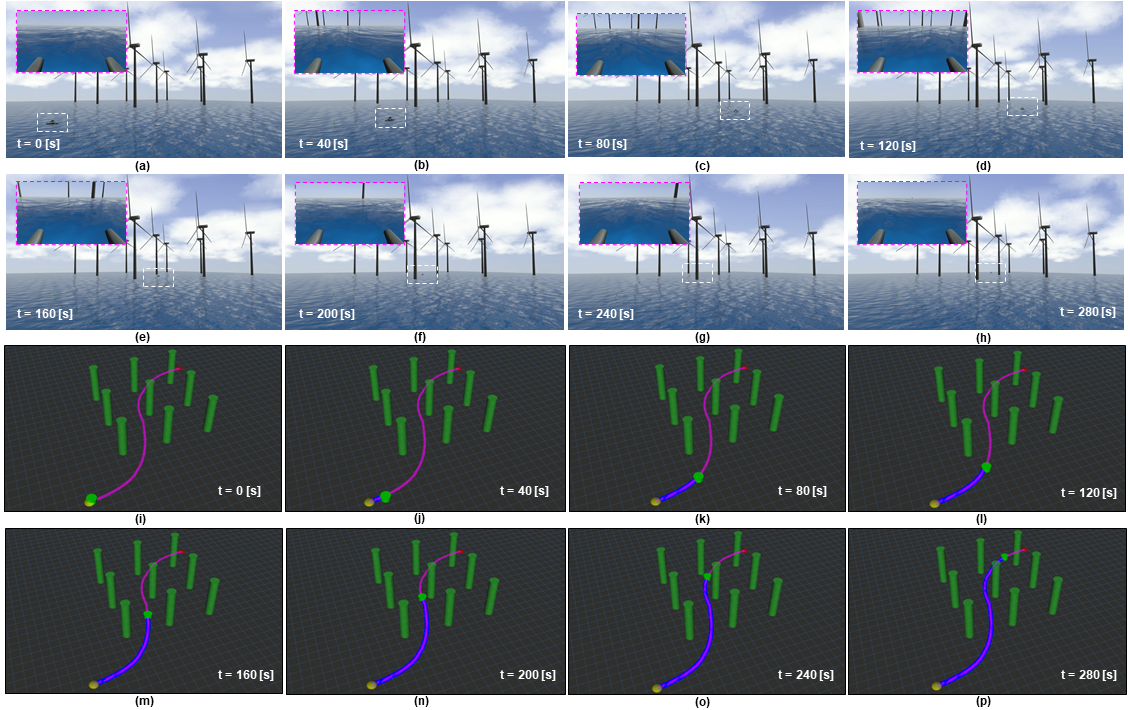}
 \centering
 \caption{The storyboards of the inspection mission in an offshore wind power generation scenario based on a path generated by MC-GPMP2*: From (a) to (h), the images demonstrate the location of the platform and the video stream from the camera mounted at the front end of the platform when the time equals 0 [s], 40 [s], 80 [s], 120 [s], 160 [s] and 200 [s], 240 [s] and 280 [s], respectively. From (i) to (\textcolor{black}{p}), the images demonstrate the corresponding motion planning problem in Rviz when the time equals 0 [s], 40 [s], 80 [s], 120 [s], 160 [s], 200 [s], 240 [s] and 280 [s], respectively.}
 \label{gpmp2_experiments} 
 \end{figure}

 \begin{figure}[t!]
 \centering
 \includegraphics[width=1.0\linewidth]{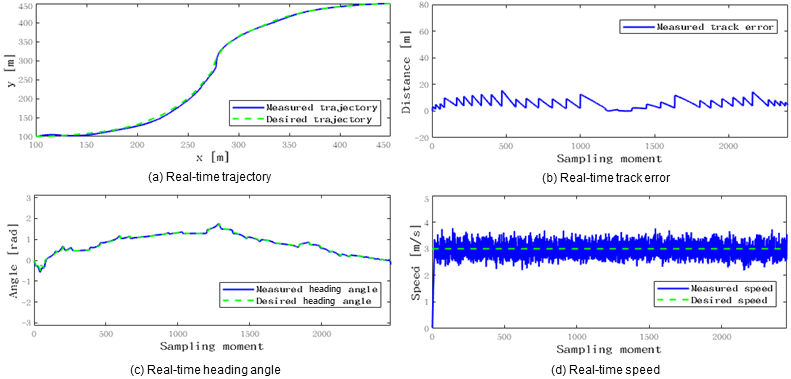}
 \centering
 \caption{Performance analysis of the inspection mission in an offshore wind power generation scenario based on a path generated by MC-GPMP2*: (a) compares the desired and measured trajectories, (b) demonstrates the real-time track error, (c) compares the desired and measured heading angles and (d) compares the desired and measured speeds.}
 \label{measurements_of_gpmp2_path} 
 \end{figure}
 
 \begin{itemize}
     \item Simulation information is inputted as: 1) a Green Buoy Model State node (/gazebo/model-states/green\_buoy) reads the location of the green buoy which represents the start point, 2) a Red Buoy Model State node (/gazebo/model-states/red\_buoy) reads the location of the red buoy which represents the goal point, 3) a WAM-V Model State node (/gazebo/model-states/wam-v) reads the current pose of the WAM-V 20 USV, 4) the Wind Turbines Model State node (/gazebo/model-states/turbines) reads the locations of the wind turbines, 5) the Ocean Currents State node (/gazebo/model-states/ocean-currents) reads the information regarding the ocean currents and 6) a WAM-V Thrusters State node (/wam-v/thrusters) reads the angle of deflection of rudders $\delta_{r}$ and thrusters' rotational speed $\omega_{t}$ in the Gazebo.
     \item This information is then transmitted and used to generate start point, goal point and obstacles in the Rviz. The motion planning algorithm then generates a desired path $\theta(t)$ with a series of waypoints $(E_{w}, N_{w})$ based on the information in Rviz.
     \item The waypoints $(E_{w}, N_{w})$ are transmitted to Gazebo and the platform begins tracking the planned path $\theta(t)$ according to the desired heading $\psi_{d}$ and the desired linear velocity $V_{d}$.
     \item The autopilot calculates and regulates the angle of deflection of rudders $\delta_{r}$ and thrusters' rotational speed $\omega_{t}$ of the platform in Gazebo in real-time according to the desired heading angle $\psi_{d}$ and linear velocity $V_{d}$. The autopilot makes the platform fulfill the motion constraint such as the pose and orientation of the desired path $\theta(t)$. It is worth noting that due to vehicle inertia, the USV would keep moving forward after reaching the target waypoint $(E_{w_{i}}, N_{w_{i}})$. In order to minimise the effects of inertia, the platform is considered to have reached the target waypoint $(E_{w_{i}}, N_{w_{i}})$ once it is inside a certain range (7 [m] in this case) of the waypoint.
     \item The platform enters the standby mode once it reaches the last target waypoint $(E_{w_{n}}, N_{w_{n}})$ of the desired path $\theta(t)$.
 \end{itemize}
 
 \subsection{Performance analysis}
 Performance analysis of the proposed autonomous navigation systems using different motion planning algorithms is detailed in Figs. \ref{measurements_of_rrt_path} and \ref{measurements_of_gpmp2_path}. In general, motion planning algorithms (such as RRT* and MC-GPMP2*) can be integrated into the proposed navigation system, within which a good trajectory tracking performance is achieved. Noticeably, as shown in Fig. \ref{rrt_experiments} and Fig. \ref{gpmp2_experiments}, MC-GPMP2* generates much smoother path than RRT*, which leads to reduced tracking error as shown in Figs. \ref{measurements_of_rrt_path} (b) and \ref{measurements_of_gpmp2_path} (b). 
 
 Improved path smoothness can also lead to less severe control inputs and potentially improve the stability of the USV. For example, by comparing the heading angles and speeds in Figs. \ref{measurements_of_rrt_path} (c), (d) and Figs. \ref{measurements_of_gpmp2_path} (c) and (d), a more gradual variation, especially in heading angle, is experienced by following the trajectory provided by MC-GPMP2* as opposed to the dramatic change between positive and negative maximum values for RRT* trajectories. Such a benefit makes the proposed MC-GPMP* a more viable solution for USVs, especially when operating in constrained areas requiring refined motion planning capability.

\section{Conclusion and future work}
 This paper introduced an improved version of the conventional GP-based motion planning algorithm (GPMP2) by further discussing the form of the likelihood function in probabilistic inference. The improved version GPMP2* extends the application scope of GPMP2 from environments with only obstacles into complex environments with a variety of environment characteristics. Further, a novel fast GP interpolation strategy with Monte-Carlo stochasticity has been added into GPMP2*, constructing another improved version named MC-GPMP2*. MC-GPMP2* can enhance the diversity in the generated path while reducing the time cost of manually tuning sampling points. Then a fully-autonomous framework has been proposed for a mainstream catamaran (WAM-V 20 USV). This framework contains an interface for any motion planner and an efficient, open-source autopilot. In four different simulations, we first demonstrated the path diversity of MC-GPMP2* and its incremental optimisation process in replanning problems. The performance of MC-GPMP2* was then compared with other mainstream motion planning algorithms such as GPMP2, A* and RRT* across a range of environments with obstacles. MC-GPMP2* generated paths with the shortest execution time, highest path smoothness and shortest path lengths in almost all cases. A competitive study was then conducted between MC-GPMP2* and a mainstream motion planning algorithm in environments with ocean currents (AFM). The results demonstrated that MC-GPMP2* delivers a better performance compared with AFM in execution time, path length and path smoothness in all the cases. Finally, we compared the performances of MC-GPMP2* and RRT* in an inspection mission based on WAM-V 20 USV and the proposed framework \textcolor{black}{in a high-fidelity virtual world}. The results further reinforced the remarkable performance of MC-GPMP2* in practical autonomous missions as well as reflected the accuracy and effectiveness of the proposed USV navigation and control framework.
 
 In terms of future work, proposed areas of focus are: 1) validating the improvement of MC-GPMP2* over other mainstream algorithms in high-dimensional environments, such as the motion planning circumstances of UUVs or robotic arms, 2) enriching the autopilot repository by adding other mainstream controllers such as back-stepping and finite-time path following, \textcolor{black}{3) automatically tuning the weight coefficients $\omega _{1}$ and $\omega _{2}$ in the objective function in (\ref{objective_function}) by using learning-based algorithms, 4) developing another motion planner that can use multiple USVs simultaneously to inspect the offshore wind power generation scenario and 5) using a digital twin for the navigation of USV so that simulation and real environment both can be benchmarked against each other.}

%\section*{Acknowledgment}


\ifCLASSOPTIONcaptionsoff
  \newpage
\fi



%\bibliographystyle{IEEEtran}

%% Bibliography
\bibliographystyle{IEEEtran}
\bibliography{IEEEtran/Paper}

\appendices
\section{\textcolor{black}{Modeling the robots and obstacles in MC-GPMP2*}}

 \begin{figure}[t!]
 \centering
 \includegraphics[width=1.0\linewidth]{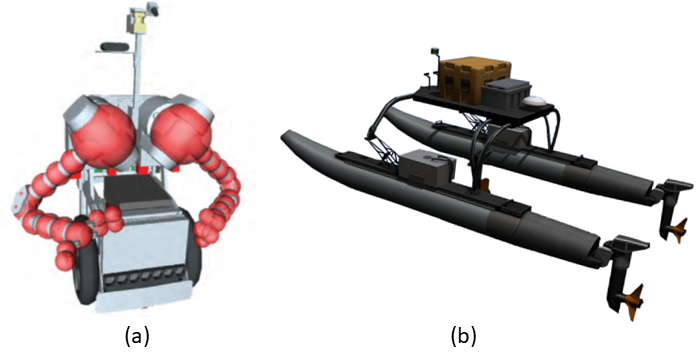}
 \centering
 \caption{\textcolor{black}{A demonstration of two different robot platforms in our problem: (a) illustrates the Herb robot \cite{zucker2013chomp} and (b) illustrates the WAM-V 20 USV \cite{vrx}.}}
 \label{robot_platforms} 
 \end{figure}

%\section{}

 \textcolor{black}{Fig. \ref{robot_platforms} provides a more intuitive perspective about the representation of different robot models in (\ref{workspace_cost_function}). To make the optimisation problem tractable, we simply view: 1) the robot model of the robotic arm as a series of spheres over the links and 2) the robot model of the catamaran as a rigid body. Consequently, (\ref{workspace_cost_function}) can be simplified as follows when we apply the catamaran as the robot platform in our motion planning problem:}
 \begin{align}
         \textcolor{black}{g_{1}(\theta_{i}) = [c(d(\theta_{i}))],}
 \label{simplified_workspace_cost_function}
 \end{align}
 
 \noindent \textcolor{black}{where $c(\cdot): \mathbb{R}^{n} \rightarrow \mathbb{R}$ is the workspace cost function that penalises the set of points $B \subset \mathbb{R}^{n}$ on the robot body when they are in or around an obstacle and $d(\cdot): \mathbb{R}^{n} \rightarrow \mathbb{R}$ is the signed distance function that calculates the signed distance of the point. Further, the signed distance function $d(\cdot)$ is defined by the following equations:}
 \begin{align}
         \textcolor{black}{d(\cdot) = D(\cdot) - \overline{D}(\cdot),}
 \label{signed_distance_field_function}
 \end{align}
 \noindent \textcolor{black}{where $D(\cdot): \mathbb{R}^{n} \rightarrow \mathbb{R}$ is the Euclidean distance transforms function. Based upon the definition in (\ref{signed_distance_field_function}), $d(\cdot)$ allows us to easily distinguish if a point is inside or outside of the obstacles. More specifically, the signed distance function: 1) generates a positive result if the point is located inside the obstacles, 2) equals to zero if the point is located on the boundaries of the obstacles and 3) generates a negative result if the point is located outside the obstacles.}
 
\section{\textcolor{black}{Density of interpolated states in GPMP2}}

\begin{figure}[t!]
 \centering
 \includegraphics[width=1.0\linewidth]{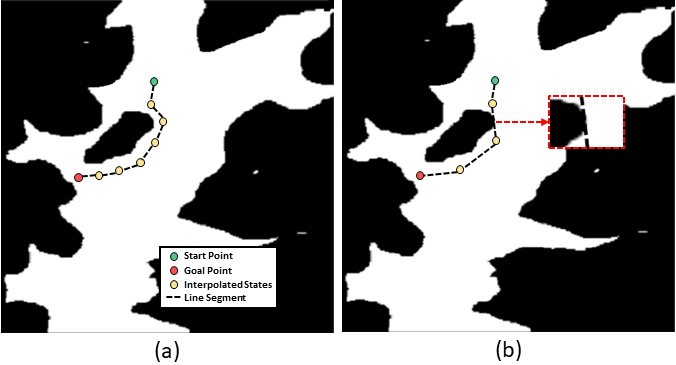}
 \centering
 \caption{\textcolor{black}{The effect of the density of interpolated states in GPMP2: (a) illustrates the trajectory generated with relatively high density on the interpolated states and (b) illustrates the trajectory generated with relatively low density on the interpolated states. In (b), the line segment between the second and third waypoints overlaps with the obstacle at the centre (a zoom-in view is provided in the red square region).}}
 \label{influence_of_waypoints_density} 
 \end{figure}

 \textcolor{black}{Based upon the information in Section. III-B in our previous research \cite{meng2022anisotropic}, we know that GPMP2 only interested in the collision-free event ($l(\theta; c_{i}=0)$). This indicates that the waypoints generated by GPMP2 are always located outside the obstacle areas to obey this rule. Whereas, the density of the interpolated states can influence the length of the line segment between two neighbour waypoints. In general, a longer line segment can increase the possibility of overlapping with obstacles as demonstrated in Fig. \ref{influence_of_waypoints_density}. To address this problem, we propose MC-GPMP2* to increase the diversity of the generated paths as well as select an appropriate number of interpolated states to ensure all the line segments between the neighbour waypoints do not overlap with any obstacle.}

\section{\textcolor{black}{USV dynamic model in ROS environment}}
 \textcolor{black}{As mentioned earlier in the introduction part of this article, we propose a fully-autonomous framework for USVs based upon the VRX simulator that is originally designed in \cite{vrx}. In this simulator, Fossen's six degrees of freedom robot-like vectorial model for marine craft \cite{fossen1999guidance, fossen1991nonlinear, fossen1995nonlinear} has been applied in Gazebo to express the dynamic model of the USV:}
 
 \begin{align}
  \textcolor{black}{\underbrace{M_{RB}\dot{v} + C_{RB}(v)v}_{\texttt{rigid\;body\;forces}} + \underbrace{M_{A}\dot{v_{r}} + C_{A}(v_{r})v_{r} + D(v_{r})v_{r}}_{\texttt{hydrodynamic\;forces}}}\\ 
  \textcolor{black}{+ \underbrace{g(\eta)}_{\texttt{hydrostatic\;forces}}}\\
  \textcolor{black}{= \tau_{propulsion} + \tau_{wind} + \tau_{waves},}
 \label{vehicle_model}
 \end{align}
 
 \noindent \textcolor{black}{where}
 
 \begin{align}
  \textcolor{black}{\eta = [x, y, z, \phi, \theta, \psi]^{T}}\\
  \textcolor{black}{v = [u, v, \omega, p, q, r]^{T},}
 \label{position_velocity_vectors}     
 \end{align}
 
 \noindent \textcolor{black}{are position and velocity vectors respectively for surge, sway, heave, roll, pitch and yaw. To be more specific, the total velocity ($v$) is the sum of an irrational water current velocity ($v_{c}$) and the vessel velocity relative to the fluid ($v_{r}$). The forces and moments due to propulsion (or the control input), wind and waves are represented as $\tau_{propulsion}$, $\tau_{wind}$ and $\tau_{waves}$. Generally, the hydrodynamic forces, hydrostatic and wave forces, wind forces and propulsion forces function on the USV simultaneously in Gazebo \cite{bingham2019toward}.}

\end{document}